\documentclass[10pt,twocolumn,letterpaper]{article}

\usepackage[pagenumbers]{wacv} 

\definecolor{wacvblue}{rgb}{0.21,0.49,0.74}
\usepackage[pagebackref,breaklinks,colorlinks,allcolors=wacvblue]{hyperref}
\usepackage{multirow}
\usepackage{graphicx}
\usepackage{tabularx}
\usepackage{nicefrac}
\usepackage[subtle]{savetrees}
\usepackage[T1]{fontenc}

\def\wacvPaperID{1364} 
\def\confName{WACV}
\def\confYear{2027}

\title{WorkIng: A Real-World and Synthetic Highway Work-Zone\\Benchmark Dataset for Autonomous Navigation}

\author{Xujun Xie$^{1}$ \quad Dominik Rößle$^{1}$ \quad Adithya Mohan$^{1}$ \quad Daniel Cremers$^{2}$ \quad Torsten Schön$^{1}$\\[2pt]
$^{1}$AImotion Bavaria, Technische Hochschule Ingolstadt, Germany\\
$^{2}$Technische Universität München, Germany\\[2pt]
{\tt\small \{xujun.xie, dominik.roessle, adithya.mohan, torsten.schoen\}@thi.de, \ cremers@tum.de}%
}

\begin{document}
\twocolumn[{%
\renewcommand\twocolumn[1][]{#1}%
\maketitle
\begin{center}
  \centering
  \includegraphics[width=\textwidth]{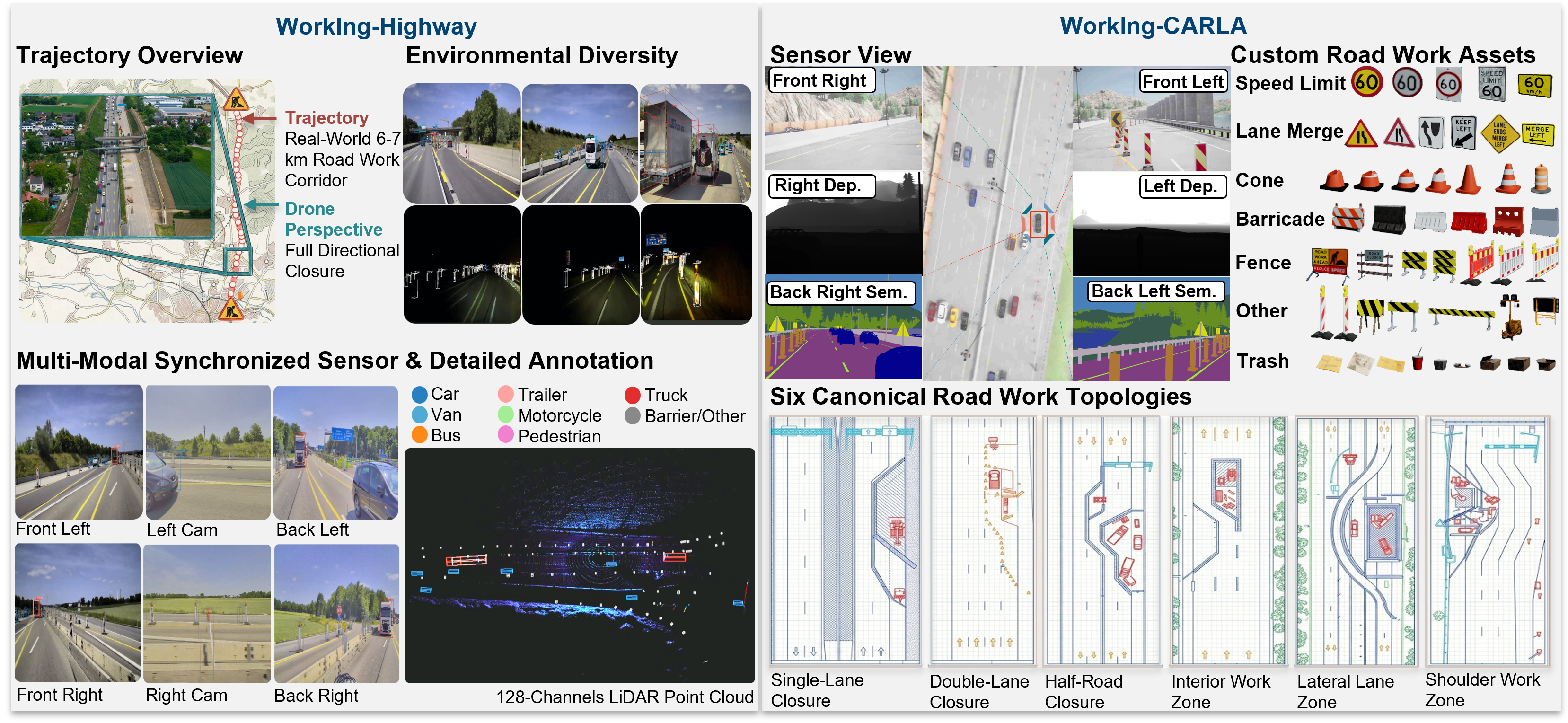}
  \captionof{figure}{\textbf{The WorkIng dataset family: A real-world and synthetic highway work-zone benchmark dataset for autonomous navigation.} \textbf{WorkIng-Highway} (left): real-world day and night driving with a $360^\circ$ camera array, 128-channel LiDAR, and dense 3D boxes. \textbf{WorkIng-CARLA} (right): the synthetic counterpart, adding depth and semantic segmentation, the custom road-work asset library, and the six canonical topologies.}
  \label{fig:dataset_overview}
\end{center}
}]

\begin{abstract}
Highway work zones are critical edge cases for autonomous vehicles (AVs), combining atypical geometries and jurisdiction-specific regulations that existing benchmarks fail to capture: real-world datasets lack $360^\circ$ multi-modal, full-scene 3D work zone annotation, and no single recorded site can span the layouts and regulatory variants needed for generalizable models. We present \textbf{WorkIng}, a dual-domain benchmark and training resource. \textbf{WorkIng-Highway} provides a $360^\circ$ multi-modal work zone benchmark with synchronized 128-channel LiDAR, six RGB cameras, and RTK-GPS across 32\,km, yielding 12{,}000+ annotated frames and 227{,}399 tracked 3D bounding boxes. \textbf{WorkIng-CARLA} complements this with procedural simulation spanning 24 regulation-aware scenarios across six canonical topologies and six international traffic standards, populated by 100+ custom 3D assets. Benchmarking representative 3D detection and end-to-end planning architectures shows that real-world data remains the primary performance driver, and that scenario-stratified synthetic augmentation improves architectures across 2D recognition, 3D detection, and trajectory planning. Resources:\\\url{https://github.com/cvims/WorkIng}.
\end{abstract}

\section{Introduction}
\label{sec:intro}

Highway work zones are a critical frontier for autonomous vehicles (AVs): standard Operational Design Domains (ODDs) assume stable lane markings, predictable signage, and uniform traffic flow~\cite{ahammed2025computer}, but work zones violate all three, forcing systems to operate under temporary regulations and non-stationary geometries~\cite{smartcities8040120, Singh2020}.

Navigating these zones is hard for two reasons. \textbf{Semantic Complexity}: atypical road topologies and construction assets are absent from standard driving ontologies~\cite{ghosh2025roadwork}. \textbf{Regulatory Variability}: standard traffic rules are suspended for local, ad-hoc ones, and the resulting regional variations (e.g., differing traffic signs and temporary lane-marker colors) can degrade perception reliability (Table~\ref{tab:regional_standards}, supplement, compares the six standards).

Existing benchmarks~\cite{Geiger2012CVPR, sun2020scalability, nuscenes, roessle2026driving, yu2020bdd100k, urbaningv2x2025} treat work zones as incidental; ROADWork~\cite{ghosh2025roadwork} is monocular RGB, lacking the LiDAR, 3D annotation, and RTK-GPS that multi-modal fusion and ego-trajectory benchmarking require; WorkZone3D~\cite{sural2026workzone3d} adds LiDAR but boxes only static, front-view objects; and synthetic benchmarks~\cite{Leaderboard2024, jia2024bench} use generic assets blind to jurisdiction (Section~\ref{sec:related_work}).

Work zone navigation therefore needs both a richly-annotated real-world dataset for multi-modal perception and a scalable simulation framework for regulatory diversity. \textbf{WorkIng} provides both under one class taxonomy, released together so the benchmark can be extended with new baselines, regions, or sensing modalities. Our contributions are:

\begin{enumerate}
    \item \textbf{WorkIng-Highway Dataset} A $360^\circ$ LiDAR-camera-RTK work-zone benchmark with tracked 3D boxes: six cameras and a 128-channel LiDAR synchronized with RTK-GPS over 19.3\,km daytime and 12.7\,km nighttime driving, yielding over $12{,}000$ annotated frames, $227{,}399$ 3D boxes, and $6{,}044$ track IDs.
    \item \textbf{WorkIng-CARLA Simulation} A procedural framework generating 24 safety-critical work zone scenarios from six canonical topologies and two interaction modes, with a \texttt{region} parameter that swaps 100+ custom 3D assets to comply with six international traffic standards.
    \item \textbf{Benchmarking and Key Insights} We evaluate representative 3D detectors (camera-only~\cite{liu2022petr}, LiDAR-only~\cite{yin2021center}, camera-LiDAR fusion~\cite{liu2022bevfusion}) and E2E planners~\cite{hu2023planning, jiang2023vad} across both domains, yielding four findings: (\textit{i})~under real-only training, illumination-invariant LiDAR gives the strongest nighttime perception, and multi-modal fusion substantially outperforms camera-only in both conditions; (\textit{ii})~the vectorized planner VAD~\cite{jiang2023vad} yields lower collision rates than the query-based UniAD~\cite{hu2023planning} in non-stationary environments, widening with prediction horizon; (\textit{iii})~real-world work zone data is the primary performance driver, while scenario-stratified synthetic augmentation adds consistent gains for \emph{all} modalities, largest for camera-dependent architectures, overturning the LiDAR-only advantage; and (\textit{iv})~this pattern extends beyond 3D detection to 2D recognition (Sec.~\ref{sec:ablation_2d}) and trajectory planning (Sec.~\ref{sec:ablation_planning}), each against an identically trained baseline.
\end{enumerate}

\section{Related Work}
\label{sec:related_work}
\subsection{Datasets for Highway Work Zone}
General-purpose benchmarks~\cite{Geiger2012CVPR, sun2020scalability, nuscenes, roessle2026driving, Argoverse2} have established the standard for urban driving with synchronized LiDAR-camera streams and dense 3D annotations, but treat work zones as incidental. The ROADWork dataset~\cite{ghosh2025roadwork} targets the domain at scale but is limited to monocular RGB, precluding 3D fusion validation; its ego-trajectory, estimated via structure-from-motion, suffers from scale ambiguity that renders precise planning benchmarking unreliable. WorkZone3D~\cite{sural2026workzone3d} adds LiDAR and 3D boxes across many U.S.\ sites, but with front-facing cameras and auto-labels for static channelizers only. \textbf{WorkIng-Highway} complements it (Table~\ref{tab:dataset_comparison}) with $360^\circ$ RTK-precise sensing, tracked boxes for traffic participants, planning evaluation, and a paired synthetic counterpart. We revisit ROADWork in Sec.~\ref{sec:ablation_2d} as the real-world target for a complementary 2D recognition evaluation of WorkIng-CARLA's synthetic work-zone assets.

\begin{table}[h]
\caption{\textbf{Comparison of autonomous driving datasets.} WZ: Work Zone coverage; Sens.: Sensor modalities (C: Camera, L: LiDAR); Cam.: Camera Coverage (FoV); Ego GT: Ego-Motion Ground Truth; WZ only: static work-zone objects; Lbl.: Label source (H: Human, A: Automatic).}
\label{tab:dataset_comparison}
\centering
\footnotesize 
\setlength{\tabcolsep}{2.5pt} 
\begin{tabular}{lcccccc}
\toprule
\textbf{Dataset} & \textbf{WZ} & \textbf{Sens.} & \textbf{Cam.} & \textbf{Ego GT} & \textbf{3D} & \textbf{Lbl.} \\ \midrule
KITTI~\cite{Geiger2012CVPR}        & Low           & C+L          & $90^\circ$   & RTK-GPS    & Yes         & H \\
nuScenes~\cite{nuscenes}          & Low           & C+L          & $360^\circ$          & RTK-GPS    & Yes         & H \\
Waymo~\cite{sun2020scalability}   & Low           & C+L          & $360^\circ$          & RTK-GPS    & Yes         & H \\
DrivIng~\cite{roessle2026driving} & Low           & C+L          & $360^\circ$          & RTK-GPS    & Yes         & H \\
BDD100K~\cite{yu2020bdd100k}       & Low           & C            & Front                & No         & No          & H \\
ROADWork~\cite{ghosh2025roadwork} & \textbf{High} & C            & Front                & Est. (Vis) & No          & H+A \\
WorkZone3D~\cite{sural2026workzone3d} & \textbf{High} & C+L       & Front                & Pose ($\sim$15\,cm) & WZ only     & A \\ \midrule
\textbf{WorkIng-Highway}          & \multirow{2}{*}{\textbf{High}} & \multirow{2}{*}{\textbf{C+L}} & \multirow{2}{*}{\textbf{360$^\circ$}} & \multirow{2}{*}{\textbf{RTK-GPS}} & \multirow{2}{*}{\textbf{Yes}} & \multirow{2}{*}{\textbf{H}} \\
\textbf{(Ours)}                   &               &              &                      &            &             &     \\ \bottomrule
\end{tabular}
\end{table}

\subsection{Simulation for Highway Work Zone}

CARLA~\cite{Dosovitskiy17} and AirSim~\cite{airsim2017fsr} provide sensory fidelity for perception training, and SOTA benchmarks~\cite{jia2024bench, Leaderboard2020, Leaderboard2024} evaluate granular driving skills, yet work zones remain a single ``default'' scenario with generic assets that ignore topological and regulatory diversity. \textbf{WorkIng-CARLA} addresses this through procedural generation grounded in official guidelines~\cite{MUTCD:2023} and parameterized across six international standards (Section~\ref{sec:carla}; Table~\ref{tab:regional_standards}, supplement).

\subsection{3D Perception and Trajectory Planning}
\textbf{3D Object Detection.} Three sensing paradigms underpin the field: \textit{LiDAR-only}~\cite{yin2021center, ZhouTuzel2018} for geometric precision; \textit{camera-only}~\cite{liu2022petr, tian2019fcos} as a cost-effective but depth-ambiguous alternative; and \textit{multi-modal fusion}~\cite{li2022bevsurvey, li2022bevformer, liu2022bevfusion} exploiting complementary sensor strengths. CenterPoint, PETR, and BEVFusion are selected as representatives across these paradigms.

\textbf{E2E Trajectory Planning.} E2E frameworks~\cite{hu2023planning, jiang2023vad} jointly optimize from perception to planning, mitigating cascading errors of modular pipelines~\cite{zhang2022mmfn}. UniAD~\cite{hu2023planning} (query-based) and VAD~\cite{jiang2023vad} (vectorized) are evaluated for trajectory stability and safety in non-stationary environments.

\section{WorkIng-Dataset}
\label{sec:dataset}

\subsection{WorkIng-Highway Dataset}
\label{sec:highway}
The dataset covers \textbf{32.0\,km} of driving (19.3\,km daytime / 12.7\,km nighttime), recorded as a round trip through a 6--7\,km construction corridor on a heavily modified section of the highway, where a full carriageway closure forces all traffic into a complex crossover configuration. The route traverses distinct work-zone phases at motorway speeds, lane reductions, lateral crossovers, varying barrier types, and barrier-lined corridors (Figure~\ref{fig:dataset_overview} top-left). Recorded about one week apart over an evolving construction site, the day and night passes capture the same corridor under differing illumination and slightly altered layouts, the temporal non-stationarity typical of real work zones. The route also includes approach and departure segments outside the active work zone; it was planned around the construction corridor, not sampled opportunistically. This yields intra-site diversity in topology, barrier type, and illumination within a single DACH campaign, complemented internationally by WorkIng-CARLA's regional asset library (Section~\ref{sec:carla}).

\textbf{Hardware Configuration.}
WorkIng-Highway was collected using a high-fidelity sensor suite comprising one LiDAR, one GPS/IMU module, and six RGB cameras, providing a complete $360^{\circ}$ surrounding view (as illustrated in Figure~\ref{fig:dataset_overview} bottom-left).
\begin{itemize}
    \item \textbf{RGB Cameras (6$\times$)}: $1920 \times 1080$ resolution @ 20\,FPS; providing $60^{\circ}$ horizontal FOV (4 cameras) and $100^{\circ}$ horizontal FOV (2 cameras).
    \item \textbf{LiDAR (1$\times$)}: 128 channels @ 20\,Hz scan rate; $360^{\circ}$ horizontal FOV, $-25^{\circ}$ to $15^{\circ}$ vertical FOV, with a range up to 240\,m (@ $\geq 10\%$ reflectivity).
    \item \textbf{GPS/IMU (1$\times$)}: RTK-corrected positioning @ 100\,Hz, achieving centimeter-level accuracy (approx. 1\,cm).
\end{itemize}

\textbf{Data Statistics.}
As detailed in Table~\ref{tab:dataset_stats}, the dataset comprises over 12,000 annotated frames, yielding a total of \textbf{227,399 bounding boxes} and \textbf{6,044 unique track IDs}. The annotations capture a diverse range of dynamic actors and critical infrastructure relevant to highway scenarios, including 70,358 cars, 35,007 trucks, and roughly 69,000 work zone asset annotations. Extended statistics and dataset samples are in the supplementary material.

\textbf{Annotation Methodology.}
Ground truth was annotated at a frequency of 5\,Hz. Each instance is represented by a 3D bounding box parameterized by its spatial center $(x, y, z)$, dimensions $(l, w, h)$, and yaw orientation $\theta$. We maintain unique, consistent tracking IDs for every object across the entire sequence to support long-term tracking and behavior analysis. The annotation process was performed by human experts, and label quality was verified through multiple rounds of visual inspection against both LiDAR point clouds and projected RGB images. Objects are categorized into nine classes (see Table~\ref{tab:dataset_stats}). Finally, to adhere to privacy regulations, all visible faces and license plates in the RGB images were anonymized using Gaussian blurring.

\begin{table}[t]
\centering
\caption{\textbf{Statistical overview of the WorkIng datasets.} Total bounding boxes (BBoxes), unique track IDs (Tracks), average instance duration (Frames), and LiDAR point density (Points) per class. n.a.\,=\,not available.}
\label{tab:dataset_stats}
\scriptsize
\setlength{\tabcolsep}{10.5pt}
\begin{tabular}{@{} ll rrrr @{}}
\toprule
\textbf{Class} & \textbf{Scenario} & \textbf{BBoxes} & \textbf{Tracks} & \textbf{Frames} & \textbf{Points} \\
\midrule
\multirow{3}{*}{Car}     & Day   & 66,197  & 1,887  & 34.9 & 481   \\
                         & Night & 4,161   & 127    & 32.8 & 447   \\
                         & CARLA & 277,874 & 7,887  & 35.2 & 572   \\ \addlinespace
\multirow{3}{*}{Van}     & Day   & 20,751  & 327    & 63.5 & 1,117 \\
                         & Night & 715     & 23     & 31.1 & 1,135 \\
                         & CARLA & 50,296  & 1,400  & 35.9 & 433   \\ \addlinespace
\multirow{3}{*}{Bus}     & Day   & 724     & 15     & 48.3 & 1,711 \\
                         & Night & n.a.    & n.a.   & n.a. & n.a.  \\
                         & CARLA & 35,660  & 723    & 49.3 & 2,343 \\ \addlinespace
\multirow{3}{*}{Truck}   & Day   & 30,596  & 583    & 52.5 & 1,334 \\
                         & Night & 4,411   & 55     & 80.2 & 1,351 \\
                         & CARLA & 44,579  & 1,220  & 36.5 & 640   \\ \addlinespace
\multirow{3}{*}{Trailer} & Day   & 26,434  & 511    & 51.7 & 1,984 \\
                         & Night & 1,924   & 48     & 40.2 & 1,979 \\
                         & CARLA & n.a.    & n.a.   & n.a. & n.a.  \\ \addlinespace
\multirow{3}{*}{Cycl.}   & Day   & 193     & 2      & 96.5 & 234   \\
                         & Night & n.a.    & n.a.   & n.a. & n.a.  \\
                         & CARLA & 34,565  & 1,211  & 28.5 & 160   \\ \addlinespace
\multirow{3}{*}{Moto.}   & Day   & 408     & 17     & 24.0 & 101   \\
                         & Night & n.a.    & n.a.   & n.a. & n.a.  \\
                         & CARLA & 74,448  & 2,105  & 35.4 & 257   \\ \addlinespace
\multirow{3}{*}{Ped.}    & Day   & 1,385   & 53     & 26.1 & 51    \\
                         & Night & 39      & 2      & 19.5 & 44    \\
                         & CARLA & 16,708  & 458    & 36.5 & 111   \\ \addlinespace
\multirow{3}{*}{Barrier} & Day   & 51,324  & 1,881  & 27.3 & 89    \\
                         & Night & 18,137  & 513    & 35.4 & 133   \\
                         & CARLA & 318,489 & 11,145 & 28.6 & 139   \\
\bottomrule
\end{tabular}
\end{table}
\subsection{WorkIng-CARLA Framework}
\label{sec:carla}

\subsubsection{Scenario Design and Coverage.}
WorkIng-CARLA provides a library of \textbf{24 distinct safety-critical scenarios}, constructed by permuting six canonical highway work zone layouts with two dynamic interaction modes. To prevent directional bias and ensure rigorous spatial evaluation, each configuration is instantiated in both left- and right-handed mirrored topologies, yielding a total of 24 unique scenarios. Implemented as parametric scenarios within the CARLA Leaderboard framework~\cite{Leaderboard2024}, these layouts are reusable templates, not a fixed set: each can be dynamically instantiated on arbitrary road segments, effectively turning any regular road into a work zone and rendering the achievable layout diversity open-ended.

\begin{table}[h]
\centering
\caption{\textbf{WorkIng-CARLA Asset Catalog.} Summary of the available 3D assets by category, visual variation (shape and color), and functional content. The library includes 100+ unique meshes and texture variations to support diverse regional standards.}
\label{tab:asset_catalog}
\footnotesize 
\setlength{\tabcolsep}{2.5pt} 

\begin{tabularx}{\columnwidth}{@{} >{\raggedright\arraybackslash}p{1.3cm} >{\raggedright\arraybackslash}p{1.3cm} >{\raggedright\arraybackslash}X >{\raggedright\arraybackslash}p{2.0cm} @{} }
\toprule
\textbf{Category} & \textbf{Asset Type} & \textbf{Visual Variations (Shape \& Color)} & \textbf{Content / Count} \\ \midrule

\textbf{Traffic Signs} & 
Speed Limit & 
\textbf{Rect:} White, Yellow (Small/Large) \newline \textbf{Circle:} White, Yellow & 
Range: 30--120\,km/h \\ \addlinespace

 & 
Warning \& Guide & 
\textbf{Rect:} White, Yellow, Blue \newline \textbf{Diamond:} Yellow \newline \textbf{Triangle:} White, Yellow & 
\textit{Directives:} Left, Right, Forward, Middle, Merge, Turn, Bend \\ \midrule

\textbf{Safety Furniture} & 
Barriers & 
Plastic (Red/White), Concrete, Steel, Wood & 
44 Variations \\ \addlinespace

 & 
Cones & 
Traffic Cones (Orange/Red), Cylinders & 
14 Variations \\ \addlinespace

 & 
Blocks & 
Roadblocks, Temporary Fencing & 
15 Variations \\ \midrule

\textbf{Markings} & 
Lane Paint & 
Lines, Directional Arrows (Yellow, Orange, White) & 
3 Types + Directional \\ \midrule

\textbf{Clutter} & 
Debris & 
Trash Bags, Boxes, Scattered Papers, Misc. Objects & 
30 Variations \\ \bottomrule

\end{tabularx}
\end{table}

\noindent \textbf{Canonical Layouts} As illustrated in Figure~\ref{fig:dataset_overview} bottom-right, six canonical work zone types are implemented, derived from standard highway work zone application patterns in FHWA guidance~\cite{MUTCD:2023}. The types overlap geometrically (e.g., a Half-Road Closure contains a Lane Closure), but each stresses a distinct challenge axis of the autonomous stack:
\begin{enumerate}
    \item \textbf{Single-Lane Closure:} The ego-vehicle recognizes lane reduction warnings and executes a safe lateral merge from a closing lane.
    \item \textbf{Multi-Lane Closure:} Sequential lane reductions require planning across multiple merge decisions under compressed spatial margins.
    \item \textbf{Half-Road Closure (Contraflow):} A high-complexity bidirectional bottleneck requiring navigation alongside oncoming traffic through a single reduced carriageway.
    \item \textbf{Interior Work Zone:} Central road obstructions within active lanes demand robust detection of median-positioned construction equipment.
    \item \textbf{Lateral Lane Shift:} Tests the agent's ability to track non-linear lane geometries (e.g., S-curves) through temporary markings without a lane reduction.
    \item \textbf{Shoulder Work Zone:} Challenges perception with lateral occlusions originating from road boundary barriers and multiple lane shifts.
\end{enumerate}

\noindent \textbf{Interaction Modes} To evaluate robustness against dynamic traffic flow, each layout is instantiated in two variants:
\begin{enumerate}
    \item \textbf{Clear Approach:} The ego-vehicle approaches the work zone with an unobstructed field of view.
    \item \textbf{Occluded Approach (Leading Vehicle):} A scripted leading vehicle directly ahead obscures signage and barriers, forcing the agent to react to sudden cut-ins or braking with limited foresight.
\end{enumerate}

\subsubsection{Regional Adaptability.}
A distinguishing feature of WorkIng-CARLA is its \textit{regional awareness}. As work zone regulations differ across jurisdictions, a configurable \texttt{region} parameter substitutes region-specific layouts, traffic sign types, and lane-marking colors at runtime.

WorkIng-CARLA supports Australia, China, Nordic, USA, Europe (Vienna Convention), and DACH standards; every sign, channelization device, and lane marking is authored against that jurisdiction's published regulation, tabulated in the supplementary material (Table~\ref{tab:regional_standards}). The 100+ custom 3D models cover regulatory signage, channelization devices, and temporary fencing with region-authentic materials (Table~\ref{tab:asset_catalog}, Figure~\ref{fig:dataset_overview} top-right); the full catalog is in the supplementary material. Releasing all configurations and assets publicly enables the community to generate jurisdiction-specific work-zone scenes without costly international field campaigns.

\subsubsection{Dataset Composition and Statistics.}
The dataset is composed of two primary categories: our novel \textit{WorkIng} work zone scenarios and standard highway maneuvers sourced from Bench2Drive~\cite{jia2024bench}.
\begin{itemize}
    \item \textbf{Work Zone Scenarios:} A total of \textbf{312} work zone episodes were generated by instantiating the 24 base layouts across diverse towns. To avoid directional bias, left- and right-side maneuvers (e.g., left vs.\ right lane merge) are generated with equal frequency for every layout.
    \item \textbf{General Highway Scenarios:} To prevent catastrophic forgetting of basic driving skills, WorkIng-CARLA includes \textbf{174} general highway episodes. These include ten distinct maneuver types: \textit{ConstructionObstacle, HighwayExit, HardBrakeRoute, MergerIntoSlowTraffic, ControlLoss, Accident, ParkedObstacle, YieldToEmergencyVehicle, HighwayCutIn,} and \textit{StaticCutIn}.
\end{itemize}

In total, the WorkIng-CARLA partition yields over \textbf{45,000 synchronized frames} collected at 5\,Hz. As detailed in Table~\ref{tab:dataset_stats}, this corresponds to \textbf{over 850{,}000 3D bounding boxes} ($277,874$ cars and $318,489$ barriers) and $26,149$ unique track IDs. Per-class statistics and samples are in the supplementary material. The asset-level CARLA taxonomy is mapped to nuScenes-compatible classes for unified cross-domain evaluation; only \textit{Trailer} is excluded, being absent from the CARLA asset catalog~\cite{carla_vehicles}.

\textbf{Data Collection Agent.} Trajectories are generated using an augmented \textbf{PDM-Lite}~\cite{Beisswenger2024PDMLite} agent extended with regulatory-compliant navigation: dynamic velocity adjustment for work zone speed limits, lane merges in response to temporary signage, and tracking cone-delineated boundaries instead of static map centerlines. The ego-vehicle sensor suite mirrors the real-world collection vehicle (Section~\ref{sec:highway}), and annotations include synchronized RGB, depth, and semantic segmentation (Figure~\ref{fig:dataset_overview} top-right).

\section{Experiment}
\label{sec:benchmarking}

\subsection{Evaluation Metrics}
The evaluation follows the protocols established by the nuScenes dataset~\cite{nuscenes} for both 3D object detection and planning tasks.

\textbf{3D Object Detection.} Performance is measured using the nuScenes evaluation protocol. Primary metrics include mean Average Precision (mAP), calculated across predefined center distance thresholds, alongside the following true-positive metrics: Average Translation Error (ATE), Average Scale Error (ASE), and Average Orientation Error (AOE). The nuScenes Detection Score (NDS, $\times100$) is the consolidated metric; lacking attributes, it omits AAE and averages $5\times$mAP with the ATE, ASE, AOE score ($\nicefrac{1}{8}$ weights). Categories are mapped onto the nuScenes taxonomy: \textit{Car} and \textit{Van} to \textit{Car}, and \textit{Cyclist} into \textit{Motorcycle} as a single two-wheeler class; all other categories remain unchanged. Classes with insufficient annotation density are excluded from evaluation per scenario to avoid unreliable AP estimates: \textit{Bus}, \textit{Motorcycle}, and \textit{Pedestrian} are excluded from Night, and \textit{Trailer} from CARLA (see Table~\ref{tab:dataset_stats}).

\textbf{Trajectory Estimation.} The spatial accuracy and safety of predicted ego-vehicle waypoints are evaluated at future time horizons of $t \in \{0.5, 1.0, 2.0\}$ seconds, as well as their temporal average. Spatial accuracy is quantified by the $L_2$ error, representing the Euclidean distance (in meters) between predicted waypoints and the ground-truth trajectory. Planning safety is assessed via the Collision Rate (\%), defined as the percentage of frames where the planned ego-vehicle trajectory intersects with the bounding box of any annotated obstacle in the scene, static or dynamic (e.g., other vehicles and pedestrians as well as work-zone barriers and cones).

\textbf{2D Asset Recognition.} For the complementary ablation in Sec.~\ref{sec:ablation_2d}, we report standard COCO-style~\cite{lin2014coco} mask Average Precision (AP), averaged over IoU thresholds $0.5$:$0.95$, on ROADWork's~\cite{ghosh2025roadwork} held-out real test set.

\subsection{Dataset Configuration}
\label{sec:dataset_config}
\noindent \textbf{WorkIng-Highway} Sequences are split into day and night subsets and each subset is segmented into 50 non-overlapping sub-sequences. Each sub-sequence is treated as an atomic unit and assigned in its entirety to a single split, so that no frames from the same sub-sequence appear in more than one split. Sub-sequences are partitioned $80\%/10\%/10\%$ (train/val/test) via stratified sampling to balance work-zone topology distribution across splits.

\textbf{WorkIng-CARLA} comprises 486 scenarios in total: training (313 = 216 work-zone + 97 general, Towns 04/06/12), validation (88=48+40), and testing (85=48+37). Validation and test sets are generated exclusively in unseen Town 13 to evaluate generalization to a novel environment. 

\subsection{Baselines}
A diverse set of representative baselines was selected covering different sensor modalities (LiDAR, Camera, Multi-modal) and E2E planning paradigms.

\textbf{3D Object Detection} Three detectors covering distinct modalities are selected: \textbf{CenterPoint}~\cite{yin2021center} (\textit{LiDAR-only}): anchor-free BEV detection, single sweep, $100\text{m}{\times}100\text{m}$ grid, voxel $(0.1,0.1,0.2)$m, 25 epochs. \textbf{PETR}~\cite{liu2022petr} (\textit{camera-only}): 3D position encoded into 2D features via FCOS3D~\cite{tian2019fcos} (V2-99-eSE), $384{\times}960$ input, 75 epochs. \textbf{BEVFusion}~\cite{liu2022bevfusion} (\textit{multi-modal}): camera–LiDAR fusion in shared BEV; Swin-Tiny~\cite{liu2021Swin} image encoder, VoxelNet~\cite{ZhouTuzel2018} LiDAR at $(0.075,0.075,0.2)$m, 50 epochs. All models are evaluated within nuScenes limits of $[\pm54\text{m}]$ on $X$ and $Y$.

\textbf{Trajectory Estimation} Two E2E transformer planners are selected: \textbf{UniAD}~\cite{hu2023planning} (\textit{query-based}): unified perception–prediction–planning; UniAD-Tiny from stage 2, 10 epochs, queue 3, voxel $(0.2,0.2,8)$m. \textbf{VAD}~\cite{jiang2023vad} (\textit{vectorized}): vectorized scene representation; VAD-Tiny, 10 epochs, voxel $(0.2,0.2,8)$m. Both are adapted from their original camera-only design to camera--LiDAR fusion, which WorkIng-Highway's surround sensing, agent tracks, and RTK trajectories support. Both omit only the HD map branch, WorkIng-Highway providing no HD map ground truth; full configs are provided in the supplementary material. All models train on 8 NVIDIA\textsuperscript{\textregistered} L40S GPUs; UniAD-Tiny (Day, 10 epochs, batch 3) requires approximately 3 days.

\textbf{2D Asset Recognition} Two representative instance segmentation architectures are evaluated for the complementary 2D ablation (Sec.~\ref{sec:ablation_2d}): \textbf{Mask R-CNN}~\cite{he2017maskrcnn} (ResNet-50-FPN) and \textbf{Mask DINO}~\cite{li2023maskdino} (DETR-based), matching ROADWork's~\cite{ghosh2025roadwork} own reported baselines for direct comparability.

\section{Benchmark Results}
\label{sec:result}
\subsection{Detection}
\label{sec:detection_results}

\subsubsection{Quantitative Results for 3D Object Detection}
\label{sec:result_detection}

Table~\ref{tab:detection_benchmark} evaluates the selected baselines under a \textbf{within-domain protocol}: Day and Night models are each trained on their respective WorkIng-Highway split and evaluated on its held-out test set; CARLA models are trained and evaluated entirely within WorkIng-CARLA. This contrasts with the ablation study (Section~\ref{sec:ablation}), where all training configurations are evaluated on the same fixed WorkIng-Highway test partition regardless of training source.

\begin{table}[h]
\centering
\caption{\textbf{3D object detection macro performance and error summary.} Evaluation of baseline architectures across Day, Night, and CARLA scenarios using standard nuScenes metrics (mAP) and true-positive error metrics (ATE, ASE, AOE). Arrows ($\uparrow/\downarrow$) indicate the direction of better performance, and \textbf{bold} text highlights the top results per scene. Class-specific mAP score matrices are detailed in the Supplemental Material. Real-world (Day/Night) results correspond to models trained on WorkIng-Highway only; CARLA results correspond to models trained on WorkIng-CARLA.}
\label{tab:detection_benchmark}
\scriptsize
\setlength{\tabcolsep}{2pt} 
\begin{tabular*}{\columnwidth}{@{\extracolsep{\fill}} ll cccc c @{}}
\toprule
\textbf{Scene} & \textbf{Model} & \textbf{ATE} $\downarrow$ & \textbf{ASE} $\downarrow$ & \textbf{AOE} $\downarrow$ & \textbf{mAP} $\uparrow$ & \textbf{NDS} $\uparrow$ \\ \midrule

\multirow{3}{*}{Day}
  & CenterPoint~\cite{yin2021center} & \textbf{0.21} & \textbf{0.18} & \textbf{0.07} & \textbf{0.76} & \textbf{79.40} \\
  & PETR~\cite{liu2022petr}          & 0.87 & 0.27 & 0.24 & 0.37 & 43.52 \\
  & BEVFusion~\cite{liu2022bevfusion} & 0.23 & 0.23 & 0.14 & 0.75 & 76.59 \\ \midrule

\multirow{3}{*}{Night}
  & CenterPoint~\cite{yin2021center} & \textbf{0.14} & \textbf{0.15} & 0.07 & \textbf{0.78} & \textbf{81.55} \\
  & PETR~\cite{liu2022petr}          & 0.87 & 0.18 & \textbf{0.06} & 0.33 & 44.04 \\
  & BEVFusion~\cite{liu2022bevfusion} & 0.22 & 0.25 & 0.20 & \textbf{0.78} & 77.71 \\ \midrule

\multirow{3}{*}{CARLA}
  & CenterPoint~\cite{yin2021center} & \textbf{0.05} & \textbf{0.15} & 0.06 & \textbf{0.97} & \textbf{94.65} \\
  & PETR~\cite{liu2022petr}          & 0.66 & 0.19 & \textbf{0.05} & 0.56 & 61.35 \\
  & BEVFusion~\cite{liu2022bevfusion} & 0.20 & 0.32 & 0.32 & 0.93 & 85.12 \\ \bottomrule
\end{tabular*}
\end{table}

\textbf{Real-World Scenarios (Day vs. Night):} LiDAR-only (CenterPoint) and multi-modal (BEVFusion) detectors consistently outperform the camera-only baseline (PETR) across both real-world splits, and CenterPoint matches or slightly exceeds BEVFusion in NDS under both Day ($79.40$ vs.\ $76.59$) and Night ($81.55$ vs.\ $77.71$) despite comparable mAP, plausibly due to the high-density 128-beam LiDAR already supplying strong geometric evidence for the large, rigid actors and barriers dominating highway work zones, leaving the camera branch to introduce calibration- and appearance-related noise that fusion does not fully suppress, an effect more pronounced at night, consistent with degraded low-light image features. PETR trails substantially in both conditions ($43.52$/$44.04$ NDS), confirming the limits of pure vision-based depth estimation. The CenterPoint--BEVFusion gap is driven by scale and orientation error (ASE, AOE), not localization (ATE); per-class score matrices are in the Supplemental Material.

Note that absolute performance scores between the Day and Night splits are not directly comparable; rare classes (Bus, Motorcycle, Pedestrian) are excluded from the Night evaluation due to insufficient annotation density (Table~\ref{tab:dataset_stats}), which naturally alters the aggregate metric.

\textbf{Simulation Environment (CARLA):} Both LiDAR-bearing detectors approach saturation (CenterPoint mAP $0.97$, BEVFusion $0.93$), reflecting controlled synthetic geometry; BEVFusion's lower NDS is driven by scale and orientation error, not missed detections. The camera gap also narrows: PETR reaches mAP $0.56$ vs.\ $0.37$ in real-world Day, consistent with rendered imagery reducing the photometric challenges limiting real-world vision-based detection, consistent with the sim-to-real gap quantified in Section~\ref{sec:ablation}.

\subsubsection{Ablation Study: Training Data Composition}
\label{sec:ablation}

To quantify the contribution of each dataset component, we evaluate 3D detection across five training configurations on the WorkIng-Highway test set (Table~\ref{tab:ablation_training}): three single-source baselines, \textbf{W} (WorkIng-Highway only), \textbf{D} (DrivIng~\cite{roessle2026driving} only), and \textbf{C} (WorkIng-CARLA only), and two incremental combinations, \textbf{W+D} and \textbf{W+D+C}. DrivIng is selected as the general-purpose baseline over alternatives like nuScenes~\cite{nuscenes} or Waymo~\cite{sun2020scalability} because its aligned 128-channel LiDAR and 6-camera $360^{\circ}$ topology avoids confounding cross-sensor adaptations, without forcing networks to resolve severe sensor mismatches. In \textbf{W+D+C}, WorkIng-CARLA is downsampled to $\nicefrac{1}{8}$ of its frames \emph{within each individual scenario}, not pooled at random, preserving complete coverage of all 24 scenarios, six topologies, and six regional standards so the mixed distribution inherits WorkIng-CARLA's full diversity.

\begin{table}[h]
\centering
\caption{\textbf{Ablation: training data composition for 3D detection.} All models evaluated on the WorkIng-Highway test set. \textbf{W}\,=\,WorkIng-Highway; \textbf{D}\,=\,DrivIng; \textbf{C}\,=\,WorkIng-CARLA. ATE/ASE/AOE: lower is better; mAP/NDS: higher is better. \textbf{Bold}: best performance per model.}
\label{tab:ablation_training}
\scriptsize
\setlength{\tabcolsep}{1.75pt}
\begin{tabular*}{\columnwidth}{@{\extracolsep{\fill}} ll ccccc ccccc @{}}
\toprule
& & \multicolumn{5}{c}{\textbf{Day}} & \multicolumn{5}{c}{\textbf{Night}} \\
\cmidrule(lr){3-7} \cmidrule(lr){8-12}
\textbf{Model} & \textbf{Train} & ATE & ASE & AOE & mAP & NDS & ATE & ASE & AOE & mAP & NDS \\
\midrule
\multirow{5}{*}{CenterPoint}
 & W     & 0.21 & \textbf{0.18 }& \textbf{0.07} & 0.76 & 79.40 & 0.14 & 0.15 & 0.07 & 0.78 & 81.55 \\
 & D     & 0.22 & 0.24 & 0.31 & 0.45 & 55.81 & 0.60 & 0.43 & 0.30 & 0.45 & 48.90 \\
 & C     & 0.50 & 0.61 & 1.07 & 0.13 & 18.37 & 0.69 & 0.67 & 0.45 & 0.25 & 30.36 \\
 & W+D   & \textbf{0.14} & \textbf{0.18} & 0.18 & 0.78 & 79.70 & 0.14 & 0.14 & 0.05 & 0.83 & 85.53 \\
 & W+D+C & \textbf{0.14} & \textbf{0.18} & 0.17 & \textbf{0.79} & \textbf{80.75} & \textbf{0.13} & \textbf{0.12} & \textbf{0.04} & \textbf{0.86} & \textbf{87.63} \\
\midrule
\multirow{5}{*}{PETR}
 & W     & 0.87 & 0.27 & 0.24 & 0.37 & 43.52 & 0.87 & \textbf{0.18} & \textbf{0.06} & 0.33 & 44.04 \\
 & D     & 1.14 & 0.39 & 0.23 & 0.11 & 22.37 & 1.09 & 0.45 & 0.59 & 0.07 & 15.17 \\
 & C     & 1.14 & 0.52 & 1.05 & 0.01 & 4.32 & 1.13 & 0.92 & 1.19 & 0.00 & 0.97 \\
 & W+D   & \textbf{0.66} & 0.19 & 0.20 & 0.37 & 47.44 & 0.61 & 0.21 & 0.11 & 0.40 & 50.70 \\
 & W+D+C & \textbf{0.66} & \textbf{0.12} & \textbf{0.07} & \textbf{0.41} & \textbf{52.27} & \textbf{0.57} & 0.26 & 0.15 & \textbf{0.44} & \textbf{52.31} \\
\midrule
\multirow{5}{*}{BEVFusion}
 & W     & 0.23 & 0.23 & 0.14 & 0.75 & 76.59 & 0.22 & 0.25 & 0.20 & 0.78 & 77.71 \\
 & D     & 0.33 & 0.42 & 1.41 & 0.08 & 15.70 & 0.65 & 0.66 & 0.91 & 0.24 & 24.50 \\
 & C     & 0.78 & 0.73 & 0.64 & 0.14 & 19.05 & 0.76 & 0.70 & 0.59 & 0.19 & 23.82 \\
 & W+D   & 0.14 & 0.18 & 0.15 & 0.79 & 81.33 & 0.16 & 0.19 & 0.14 & 0.82 & 82.76 \\
 & W+D+C & \textbf{0.12} & \textbf{0.13} & \textbf{0.08} & \textbf{0.83} & \textbf{84.98} & \textbf{0.15} & \textbf{0.14} & \textbf{0.11} & \textbf{0.88} & \textbf{87.69} \\
\bottomrule
\end{tabular*}
\end{table}

\textbf{General-purpose training does not generalize to work zones.} Models trained exclusively on DrivIng perform poorly across all architectures (e.g., CenterPoint NDS: $55.81$ Day, $48.90$ Night), confirming that domain-specific work zone data is essential for reliable detection.

\textbf{Synthetic-only training does not transfer to real work zones.} Models trained exclusively on WorkIng-CARLA achieve NDS from $0.97$ (PETR Night) to $30.36$ (CenterPoint Night) on real WorkIng-Highway sequences, far below any configuration including real data, confirming synthetic data alone cannot substitute for real-world collection, and motivating its role as a \textit{complementary} resource.

\textbf{Real-world data is the primary performance driver; synthetic augmentation benefits every modality.} Adding WorkIng-Highway to the DrivIng baseline closes the work zone performance gap across all three architectures. Adding WorkIng-CARLA on top further improves \emph{every} architecture in \emph{both} lighting conditions (Table~\ref{tab:ablation_training}), with the largest gains accruing to camera-branch architectures (mean $\Delta$NDS: BEVFusion $+4.29$, PETR $+3.22$, CenterPoint $+1.58$), inverting the intuition that synthetic imagery is primarily a liability for appearance-dependent models.

\textbf{Synthetic augmentation unlocks the camera branch in multi-modal fusion.} Under real-only training, LiDAR-only CenterPoint matches or exceeds multi-modal BEVFusion (Section~\ref{sec:result_detection}); this ordering reverses once WorkIng-CARLA is included, with BEVFusion surpassing CenterPoint by $4.23$ NDS during the day ($84.98$ vs.\ $80.75$) and matching it at night ($87.69$ vs.\ $87.63$). This is consistent with a data-starved, not uninformative, camera stream, though the ablation does not isolate this mechanism.

\subsubsection{Ablation Study: 2D Work-Zone Asset Recognition}
\label{sec:ablation_2d}

As no other-region work-zone dataset offers compatible full-scene 3D labels (WorkZone3D~\cite{sural2026workzone3d} annotates only cones, barrels, and channelizers in 3D), we instead evaluate whether WorkIng-CARLA's synthetic assets transfer to real work-zone imagery, using ROADWork's~\cite{ghosh2025roadwork} 2D instance segmentation on real U.S.\ data. This measures aggregate transfer of the asset library; whether \emph{region-matched} assets transfer better than region-mismatched ones is a separate question we do not test.

We export WorkIng-CARLA's 3D annotations as COCO-format 2D instance masks under ROADWork's exact taxonomy and evaluate the baselines introduced in Sec.~\ref{sec:benchmarking}, alongside three zero-shot open-vocabulary foundation models reported by ROADWork~\cite{ghosh2025roadwork}, across three training configurations analogous to the trajectory sim-to-real ablation (Section~\ref{sec:ablation_planning}): \textbf{C} (WorkIng-CARLA only), \textbf{R} (ROADWork only), and \textbf{C+R} (combined), evaluated on ROADWork's held-out real test set. Full method is described in the Supplementary Material, Section~\ref{supp:2d_asset_detection}.

\begin{table}[h]
\centering
\caption{\textbf{2D asset-recognition ablation.} Mask AP (\%) on ROADWork's real test set, COCO protocol. Top block: zero-shot foundation models, values as reported by ROADWork~\cite{ghosh2025roadwork} (not reproduced here). Remaining blocks: this work, all three configurations trained under a single identical recipe per architecture. \textbf{C}\,=\,WorkIng-CARLA only; \textbf{R}\,=\,ROADWork only; \textbf{C+R}\,=\,combined. \textbf{Bold}: best per architecture.}
\label{tab:ablation_2d}
\scriptsize
\setlength{\tabcolsep}{3pt}
\begin{tabular*}{\columnwidth}{@{\extracolsep{\fill}} l cccccc @{}}
\toprule
\textbf{Model} & AP & AP\textsubscript{50} & AP\textsubscript{75} & AP\textsubscript{S} & AP\textsubscript{M} & AP\textsubscript{L} \\
\midrule
\multicolumn{7}{@{}l}{\textit{Zero-shot foundation models~\cite{ghosh2025roadwork}}} \\
OpenSeeD~\cite{zhang2023openseed} & 2.9 & 5.8 & 2.5 & 1.3 & 2.8 & 4.0 \\
Detic~\cite{zhou2022detic}        & 4.2 & 6.3 & 4.6 & 2.1 & 5.5 & 6.0 \\
OMG-LLaVA~\cite{zhang2024omgllava} & 0.2 & 0.3 & 0.2 & 0.0 & 0.1 & 0.8 \\
\midrule
\multicolumn{7}{@{}l}{\textit{Train: C (WorkIng-CARLA only)}} \\
Mask R-CNN~\cite{he2017maskrcnn}  & 4.13 & 5.07 & 4.23 & 2.74 & 3.48 & 5.12 \\
Mask DINO~\cite{li2023maskdino}   & 5.65 & 8.82 & 5.05 & 2.49 & 4.46 & 7.44 \\
\midrule
\multicolumn{7}{@{}l}{\textit{Train: R (ROADWork only)}} \\
Mask R-CNN~\cite{he2017maskrcnn}  & 30.23 & 49.30 & 32.41 & 17.09 & 32.78 & \textbf{42.60} \\
Mask DINO~\cite{li2023maskdino}   & 38.01 & 55.70 & 41.76 & 19.20 & 38.12 & \textbf{51.27} \\
\midrule
\multicolumn{7}{@{}l}{\textit{Train: C+R (combined)}} \\
Mask R-CNN~\cite{he2017maskrcnn}  & \textbf{35.56} & \textbf{50.65} & \textbf{34.42} & \textbf{19.58} & \textbf{33.74} & 39.39 \\
Mask DINO~\cite{li2023maskdino}   & \textbf{43.15} & \textbf{59.73} & \textbf{44.01} & \textbf{21.02} & \textbf{43.47} & 47.54 \\
\bottomrule
\end{tabular*}
\end{table}

\textbf{Synthetic augmentation improves 2D recognition as it does 3D, and WorkIng-CARLA's asset library alone provides more task-relevant signal than web-scale open-vocabulary pretraining.} Trained on \textbf{C} alone, with no real imagery, Mask DINO reaches $5.65$ AP, above the strongest zero-shot foundation model ROADWork reports (Detic, $4.2$); \textbf{C+R} then improves AP over \textbf{R}-only by $+17.6\%$ (Mask R-CNN) and $+13.5\%$ (Mask DINO) relative, extending the same synthetic-augmentation benefit observed for 3D detection (Section~\ref{sec:ablation}). Because ROADWork releases no Mask DINO configuration, we train \textbf{C}, \textbf{R}, and \textbf{C+R} under a single identical recipe per architecture; our \textbf{R}-only baselines reach $30.23$ and $38.01$ AP against the $29.9$ and $36.4$ reported in~\cite{ghosh2025roadwork}, so the \textbf{C+R} gain is measured against a like-for-like baseline. Full discussion is in the Supplementary Material, Section~\ref{supp:2d_asset_detection}.

\subsection{Trajectory Planning}
\label{sec:trajectory_results}

\subsubsection{Quantitative Results for Trajectory Estimation}
\label{sec:result_planning}

Table~\ref{tab:planning_benchmark} summarizes the open-loop trajectory forecasting performance of UniAD and VAD across Day and Night, following the same within-domain protocol as the detection benchmark (Section~\ref{sec:result_detection}). WorkIng-CARLA's closed-loop planning results are reported separately in Section~\ref{sec:closed_loop}, and cross-domain sim-to-real transfer for trajectory planning in Section~\ref{sec:ablation_planning}.

\begin{table}[h]
\centering
\caption{\textbf{Quantitative comparison of planning performance across different scenes.} The L2 displacement error (in meters) and the collision rate (in \%) are evaluated at future time horizons of 0.5s, 1.0s, and 2.0s, along with their average. For both metrics, lower values are better ($\downarrow$). \textbf{Bold} values denote the best performance in each scenario.}
\label{tab:planning_benchmark}
\scriptsize 
\setlength{\tabcolsep}{2pt} 

\begin{tabularx}{\columnwidth}{@{} l >{\raggedright\arraybackslash}p{1.7cm} *{8}{>{\centering\arraybackslash}X} @{}}
\toprule
\multirow{2}{*}{\textbf{Scene}} & \multirow{2}{*}{\textbf{Model}} & \multicolumn{4}{c}{\textbf{L2 Error (m)} $\downarrow$} & \multicolumn{4}{c}{\textbf{Collision Rate (\%)} $\downarrow$} \\ \cmidrule(lr){3-6} \cmidrule(lr){7-10}
 & & 0.5s & 1.0s & 2.0s & Avg. & 0.5s & 1.0s & 2.0s & Avg. \\ \midrule

\multirow{2}{*}{Day}
  & UniAD\_tiny \cite{hu2023planning} & 0.23 & 0.55 & 1.45 & 0.74 & \textbf{0.11} & \textbf{0.33} & 5.97 & 2.14 \\
  & VAD\_tiny \cite{jiang2023vad}    & \textbf{0.18} & \textbf{0.43} & \textbf{1.25} & \textbf{0.62} & 0.14 & 0.34 & \textbf{5.13} & \textbf{1.87} \\ \midrule

\multirow{2}{*}{Night}
  & UniAD\_tiny \cite{hu2023planning} & 0.37 & \textbf{0.70} & 1.63 & \textbf{0.90} & \textbf{0.85} & \textbf{0.88} & 4.84 & 2.19 \\
  & VAD\_tiny \cite{jiang2023vad}    & \textbf{0.35} & 0.86 & \textbf{1.60} & 0.94 & \textbf{0.85} & \textbf{0.88} & \textbf{4.76} & \textbf{2.16} \\ \bottomrule

\end{tabularx}
\end{table}

\textbf{Overall Performance and Temporal Horizons:}
 VAD-Tiny achieves the lower average collision rate in both Day and Night and the lower average $L_2$ error during the day ($0.62$ vs.\ $0.74$), while UniAD-Tiny is marginally ahead on average $L_2$ at night ($0.90$ vs.\ $0.94$). The advantage grows with horizon: at 2.0\,s VAD-Tiny reduces collision rate by $14\%$ during the day ($5.13\%$ vs.\ $5.97\%$), whereas at 0.5--1.0\,s the two are comparable and UniAD-Tiny occasionally leads. VAD-Tiny's advantage is therefore horizon-dependent.

\textbf{Environmental Impacts (Day vs. Night):}
Night scenarios increase average $L_2$ error for both models (VAD: $0.62$\,m $\to$ $0.94$\,m), reflecting degraded upstream perception under low light, and raise the average collision rate slightly (VAD: $1.87\%\to2.16\%$).

\subsubsection{Closed-Loop Evaluation}
\label{sec:closed_loop}

To directly evaluate whether trajectory accuracy under open-loop replay translates to actual driving competence, we evaluate both E2E planners, trained on WorkIng-CARLA only, under the CARLA Leaderboard 2.0~\cite{Leaderboard2024} closed-loop protocol on the same WorkIng-CARLA test scenarios (Section~\ref{sec:dataset_config}). Unlike open-loop replay, the simulator reacts to the agent's own predicted trajectory instead of a logged reference. We report $L_2$ error and Collision Rate under this reactive rollout, alongside the standard CARLA Leaderboard Driving Score (DS), decomposed into Route Completion (RC) and Infraction Penalty (IP)~\cite{Prakash2021CVPR, Chitta2023PAMI}. Closed-loop evaluation is only possible within CARLA -- WorkIng-Highway's real-world sequences are logged sensor data, not a reactive simulator.

\begin{table}[h]
\centering
\caption{\textbf{Closed-loop CARLA Leaderboard evaluation.} $L_2$ error (m) and Collision Rate (\%) measured under reactive closed-loop rollout, alongside Driving Score (DS, composed), Route Completion (RC), and Infraction Penalty (IP). Higher is better for DS/RC/IP; lower is better for $L_2$/Collision Rate. \textbf{Bold}: best per column.}
\label{tab:closed_loop}
\scriptsize
\setlength{\tabcolsep}{2.5pt}
\begin{tabular*}{\columnwidth}{@{\extracolsep{\fill}} l cccc cccc ccc @{}}
\toprule
& \multicolumn{4}{c}{\textbf{L2 Error (m)} $\downarrow$} & \multicolumn{4}{c}{\textbf{Collision Rate (\%)} $\downarrow$} & \multicolumn{3}{c}{\textbf{LB2.0 Score} $\uparrow$} \\
\cmidrule(lr){2-5} \cmidrule(lr){6-9} \cmidrule(lr){10-12}
\textbf{Model} & 0.5s & 1.0s & 2.0s & Avg. & 0.5s & 1.0s & 2.0s & Avg. & DS & RC & IP \\
\midrule
UniAD\_tiny~\cite{hu2023planning} & \textbf{0.23} & \textbf{0.48} & 1.34 & 0.68 & 0.23 & 0.75 & 2.84 & 1.27 & 0.63 & \textbf{0.80} & 0.79 \\
VAD\_tiny~\cite{jiang2023vad}    & 0.25 & 0.54 & \textbf{1.19} & \textbf{0.66} & \textbf{0.08} & \textbf{0.54} & \textbf{1.96} & \textbf{0.86} & \textbf{0.65} & 0.76 & \textbf{0.85} \\
\bottomrule
\end{tabular*}
\end{table}

\textbf{Closed-loop performance is too close to separate the two planners.} Under reactive rollout the two are comparable: the composed Driving Score differs by $0.02$ ($0.65$ vs.\ $0.63$) on a single run over 85 routes, which we do not treat as a ranking. The decomposition is nonetheless informative: UniAD-Tiny completes more of the route (Route Completion $0.80$ vs.\ $0.76$) by driving more aggressively, while VAD-Tiny commits proportionally fewer infractions (Infraction Penalty $0.85$ vs.\ $0.79$). This progress-versus-safety trade-off cannot be expressed under open-loop replay, where route progress is fixed by the logged reference; surfacing it is what WorkIng-CARLA's closed-loop protocol adds beyond the open-loop metrics.

\subsubsection{Ablation Study: Trajectory Sim2Real}
\label{sec:ablation_planning}

Mirroring the 3D detection ablation (Section~\ref{sec:ablation}), we evaluate trajectory planning across three training configurations on the WorkIng-Highway test set: \textbf{W} (WorkIng-Highway only), \textbf{C} (WorkIng-CARLA only), and \textbf{W+C} (combined), using the same scenario-stratified sampling of WorkIng-CARLA described in Section~\ref{sec:ablation}.

\textbf{Synthetic-only training does not transfer to real trajectory planning.} Models trained exclusively on WorkIng-CARLA achieve $L_2$ error of $7$--$9$\,m and collision rates of $5.5$--$7\%$ on real WorkIng-Highway sequences, an order of magnitude worse than any configuration including real data -- the same sim-to-real gap already established for 3D detection (Section~\ref{sec:ablation}).

\begin{table}[t]
\centering
\caption{\textbf{Ablation: training data composition for trajectory planning.} $L_2$ error (m) and Collision Rate (\%), averaged over 0.5/1.0/2.0s horizons, on the WorkIng-Highway test set. \textbf{W}\,=\,WorkIng-Highway; \textbf{C}\,=\,WorkIng-CARLA; D/N\,=\,Day/Night. Lower is better for both metrics. \textbf{Bold}: best per model/scene.}
\label{tab:ablation_planning}
\scriptsize
\setlength{\tabcolsep}{2.5pt}
\renewcommand{\arraystretch}{0.9}
\begin{tabular*}{\columnwidth}{@{\extracolsep{\fill}} l cccc cccc @{}}
\toprule
& \multicolumn{4}{c}{\textbf{UniAD\_tiny}} & \multicolumn{4}{c}{\textbf{VAD\_tiny}} \\
\cmidrule(lr){2-5} \cmidrule(lr){6-9}
& \multicolumn{2}{c}{$L_2$} & \multicolumn{2}{c}{Coll.\%} & \multicolumn{2}{c}{$L_2$} & \multicolumn{2}{c}{Coll.\%} \\
\textbf{Train} & D & N & D & N & D & N & D & N \\
\midrule
W   & 0.74 & 0.90 & 2.14 & 2.19 & 0.62 & 0.94 & 1.87 & 2.16 \\
C   & 7.57 & 8.13 & 6.98 & 5.51 & 8.32 & 9.33 & 6.78 & 5.82 \\
W+C & \textbf{0.66} & \textbf{0.84} & \textbf{1.17} & \textbf{1.65} & \textbf{0.53} & \textbf{0.83} & \textbf{1.09} & \textbf{1.78} \\
\bottomrule
\end{tabular*}
\renewcommand{\arraystretch}{1}
\end{table}

\textbf{Synthetic augmentation improves trajectory planning across every model, scene, and metric.} Adding WorkIng-CARLA to WorkIng-Highway training improves both $L_2$ error and collision rate in all four (model, scene) combinations, with no exceptions: $L_2$ error falls by $7$--$15\%$ (e.g.\ VAD-Tiny Day: $0.62\to0.53$), while collision rate falls by $18$--$45\%$ (e.g.\ UniAD-Tiny Day: $2.14\%\to1.17\%$) -- a substantially larger relative gain on safety than on raw trajectory accuracy. This replicates the 3D detection pattern on a second task.

\section{Conclusion}
\label{sec:conclusion}

\textbf{WorkIng} pairs a real-world $360^\circ$ multi-modal work-zone benchmark with a regulation-parameterized synthetic counterpart under a shared taxonomy, benchmarked across 3D detection, 2D recognition, and open- and closed-loop trajectory planning (Section~\ref{sec:result}).

\textbf{Limitations and Future Work.} WorkIng-Highway's single corridor is its principal limitation. Work zones are transient, and recording and publicly releasing active German work zones requires construction-authority approval and anonymization of on-site workers; we therefore prioritized depth over breadth, choosing one corridor of maximal internal diversity. Real and synthetic data remain complementary, not interchangeable: WorkIng-CARLA-only models transfer poorly to real sequences (Table~\ref{tab:ablation_training}). Section~\ref{sec:ablation_2d} shows synthetic assets improve recognition on real U.S.\ imagery, but region-matched versus region-mismatched transfer, and cross-regional 3D detection and planning, remain untested; multi-site, multi-jurisdiction collection is future work.

\section*{Acknowledgments}
This work was funded by the iEXODDUS project (Grant Agreement No. 101146091) and partly by the Hightech Agenda Bayern.

{
    \small
    \bibliographystyle{ieeenat_fullname}
    \bibliography{main}
}
\clearpage
\section{Supplemental Material}
This document provides extended technical specifications, quantitative analyses, and visual catalogs accompanying the main manuscript. The supplemental material is structured into five primary divisions corresponding to the core data domains and the extended benchmark analyses:
\begin{itemize}
    \item \textbf{Dataset Design Rationale and Scope:} Answers two dataset-design questions raised during review: the single-corridor scope of WorkIng-Highway and the per-split class exclusions.
    \item \textbf{WorkIng-Highway:} Details the real-world sensor configuration, precise annotation taxonomy, and spatial distribution statistics.
    \item \textbf{WorkIng-CARLA:} Presents the synthetic work-zone asset library and the scenario generation distributions.
    \item \textbf{Benchmark: Extended Analysis:} Details a per-class decomposition of the overall mean Average Precision (mAP) metrics and offers a qualitative review of representative failure and success cases for each sensor modality.
    \item \textbf{2D Work-Zone Asset Detection:} Presents the method and extended results for the complementary 2D asset-recognition ablation (Section~\ref{sec:ablation_2d}).
\end{itemize}

\subsection{Dataset Design Rationale and Scope}
\label{supp:qa}

This section addresses two dataset-design questions raised during review that are best answered directly.

\noindent\textbf{Why does WorkIng-Highway cover one corridor instead of several work zones?} Highway work zones are transient by nature: a site's lane configuration, barrier placement, and signage can change from one week to the next as construction progresses, so a fixed multi-site sampling plan risks going stale before it can be executed. Access is also gated by construction-authority approval, a slower and more constrained process than typical urban-street permitting. Given these constraints, we prioritized depth over breadth: a single corridor selected specifically for having the highest internal diversity of work-zone configurations we could find (multiple lane shifts, varying barrier types, transitioning lane configurations, matched day and night passes), instead of shallow coverage spread across many sites. To our knowledge, no prior dataset provides synchronized LiDAR-camera-RTK ground truth for highway work zones at all; WorkIng-CARLA's six-standard synthetic asset library (Section~\ref{sec:carla}) is our approach to supplying the regulatory and topological diversity that a single real corridor cannot, and Section~\ref{sec:ablation_2d} evaluates whether those assets transfer to real imagery from a jurisdiction we never recorded.

\noindent\textbf{Why do the classes excluded from evaluation differ across the Day, Night, and CARLA splits (Section~\ref{sec:benchmarking})?} Construction crews are not present uniformly along the length of a work zone or across every recording pass: some segments and time windows have active workers and work vehicles present, others do not, reflecting how real construction sites operate; it is not an artifact of the recording protocol. This makes an evenly balanced class distribution across splits unachievable without either fabricating instances or discarding real recorded segments, neither of which is desirable. We instead retain all recorded data and exclude, per scenario, only the classes whose annotation density is too low in that scenario to support a reliable AP estimate (Section~\ref{sec:benchmarking}); the full per-class instance counts underlying this decision are given in Table~\ref{tab:dataset_stats}.

\subsection{WorkIng-Highway}

\subsubsection{Sensor Setup}
\begin{figure}[h]
    \centering
    \includegraphics[width=\linewidth]{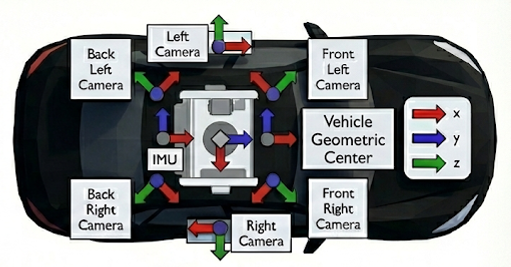}
    \caption{\textbf{Overhead view of the sensor layout.} Labels denote the positions of six Surround-View Cameras and an IMU, with a legend defining the vehicle-fixed coordinate frame $(x, y, z)$.}
    \label{fig:sensor_placement}
\end{figure}

As illustrated in Figure~\ref{fig:sensor_placement}, a right-handed, vehicle-fixed coordinate system $(x, y, z)$ is established with the origin at the Vehicle Geometric Center. The sensors are positioned to provide a continuous $360^{\circ}$ horizontal FOV while maintaining overlap for multi-view consistency.

\noindent \textbf{Corner Peripherals:} Four cameras with a $60^{\circ}$ horizontal FOV are located at the vehicle corners (Front-Left, Front-Right, Back-Left, and Back-Right) for long-range perception of approaching and receding traffic.

\noindent \textbf{Lateral Monitoring:} Two cameras with a $100^{\circ}$ horizontal FOV are mounted at the lateral mid-points of the vehicle, providing broader coverage of the immediate proximity.

\noindent \textbf{Central Reference Unit:} The LiDAR is located on the vehicle roof; GPS/IMU modules are mounted inside the vehicle.

\subsubsection{Annotation Details}
Objects are annotated with specific attributes based on their semantic category. These attributes describe the physical state, operational role, or subtype of the detected instances. Table~\ref{tab:attributes} illustrates the mapping between defined object categories and their corresponding attribute types. For instance, the \textit{Pedestrian} category is further refined by posture and activity attributes (e.g., standing, walking, sitting), while vehicle categories like \textit{Bus} and \textit{Truck} include structural attributes (e.g., bendy, rigid) and operational roles (e.g., public transport, emergency).

\begin{table}[h]
  \footnotesize
  \centering
  \caption{Mapping of object categories to specific attribute types within the annotation scheme.}
  \label{tab:attributes}
  \begin{tabular}{ll}
    \toprule
    \textbf{Attributes} & \textbf{Categories} \\
    \midrule
    emergency, regular, public transport & Car, Van, Bus, Truck \\
    car trailer, truck trailer, cyclist trailer & Trailer \\
    standing, walking, sitting & Pedestrian \\
    bendy, rigid & Bus, Truck \\
    None & Motorcycle, Bicycle, Barrier \\
    \bottomrule
  \end{tabular}
\end{table}

\subsubsection{Extended Dataset Statistics}
\begin{figure*}[t]
     \centering
     \begin{subfigure}[b]{0.32\textwidth}
         \centering
         \includegraphics[width=\textwidth]{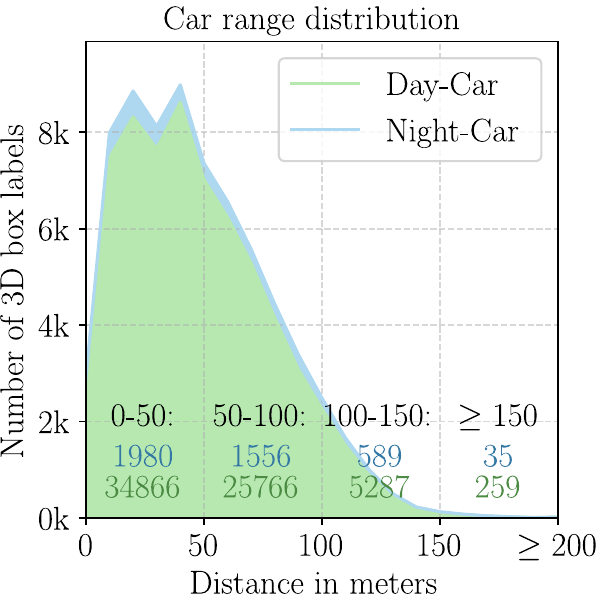}
     \end{subfigure}
     \hfill
     \begin{subfigure}[b]{0.32\textwidth}
         \centering
         \includegraphics[width=\textwidth]{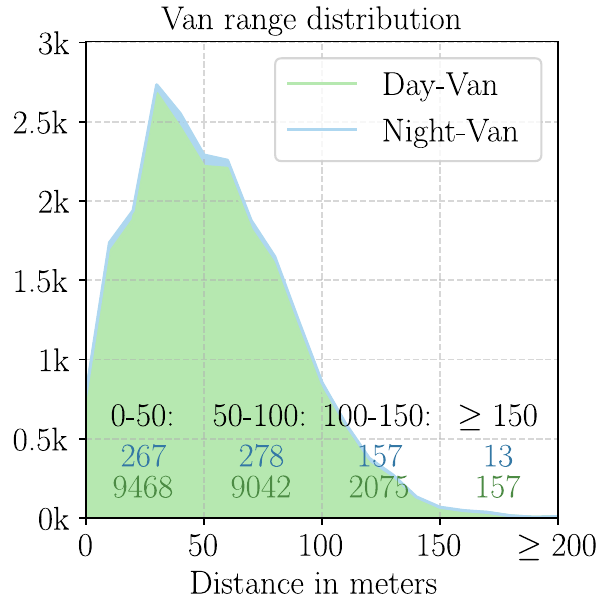}
     \end{subfigure}
     \hfill
     \begin{subfigure}[b]{0.32\textwidth}
         \centering
         \includegraphics[width=\textwidth]{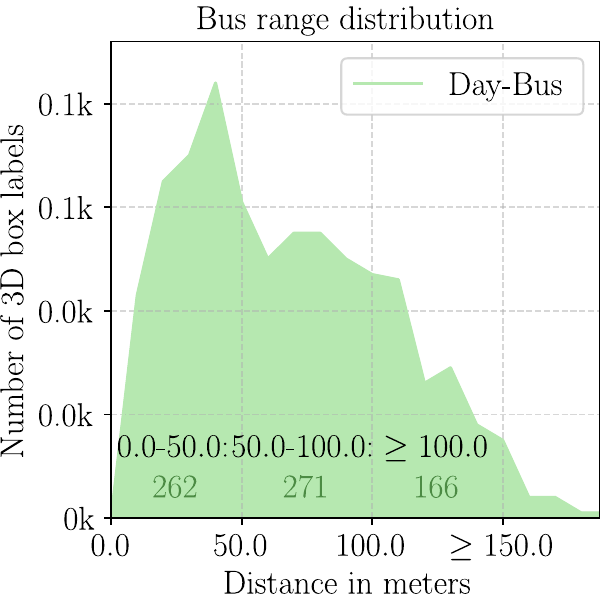}
     \end{subfigure}

     \vspace{1em} 

     \begin{subfigure}[b]{0.32\textwidth}
         \centering
         \includegraphics[width=\textwidth]{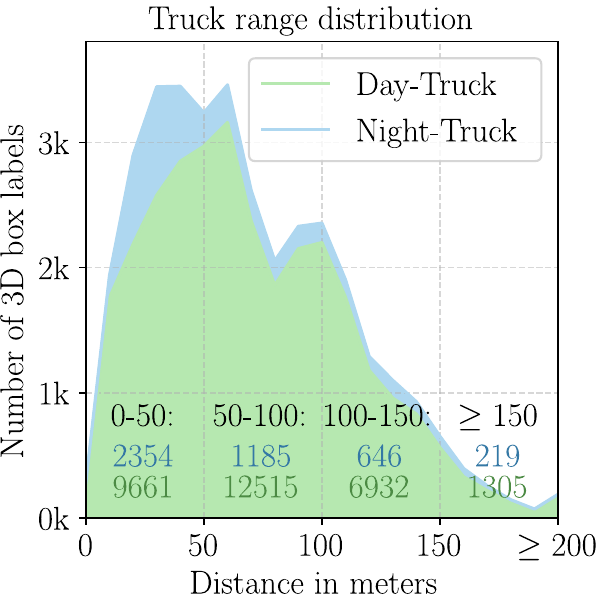}
     \end{subfigure}
     \hfill
     \begin{subfigure}[b]{0.32\textwidth}
         \centering
         \includegraphics[width=\textwidth]{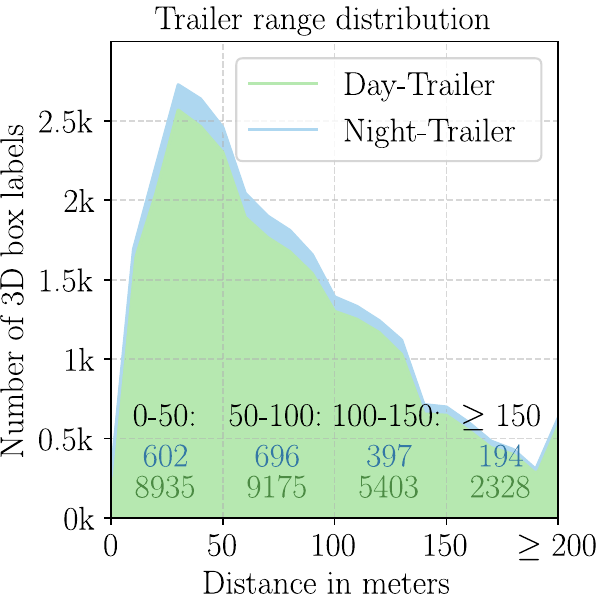}
     \end{subfigure}
     \hfill
     \begin{subfigure}[b]{0.32\textwidth}
         \centering
         \includegraphics[width=\textwidth]{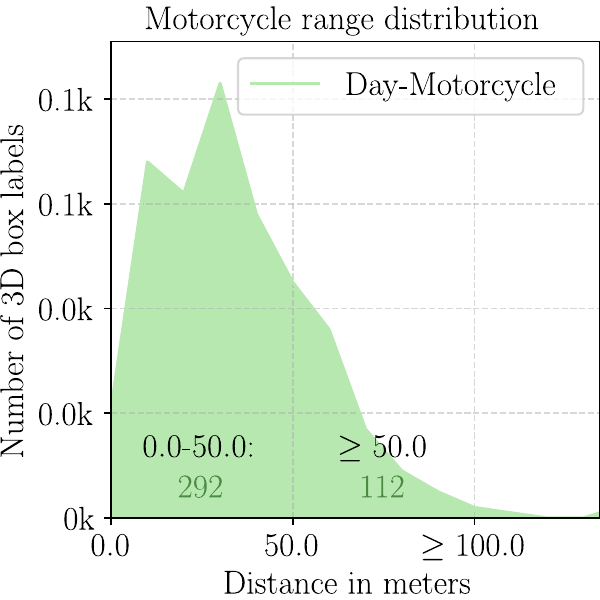}
     \end{subfigure}

     \vspace{1em}

     \begin{subfigure}[b]{0.32\textwidth}
         \centering
         \includegraphics[width=\textwidth]{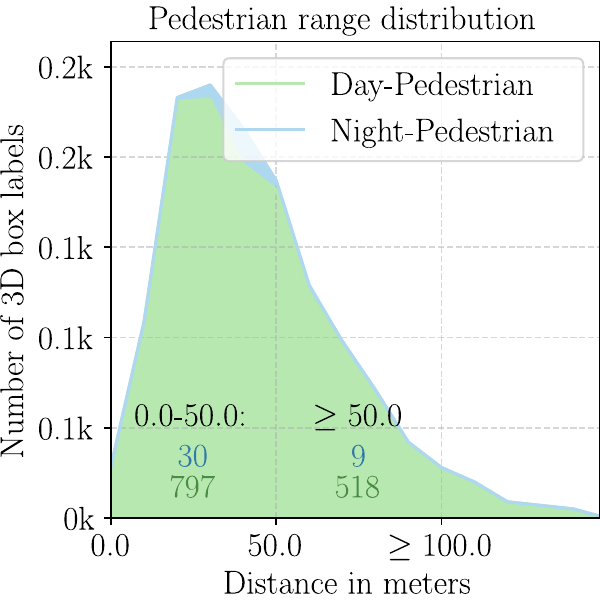}
     \end{subfigure}
     \hfill
     \begin{subfigure}[b]{0.32\textwidth}
         \centering
         \includegraphics[width=\textwidth]{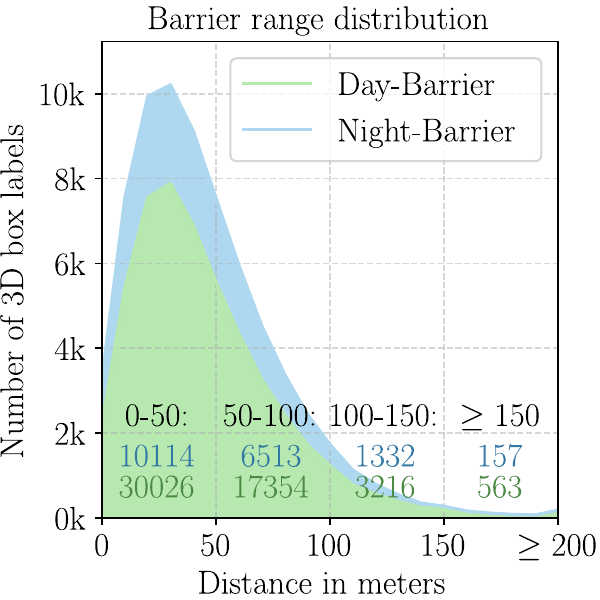}
     \end{subfigure}
     \hfill
     \begin{subfigure}[b]{0.32\textwidth}
         \centering
         \includegraphics[width=\textwidth]{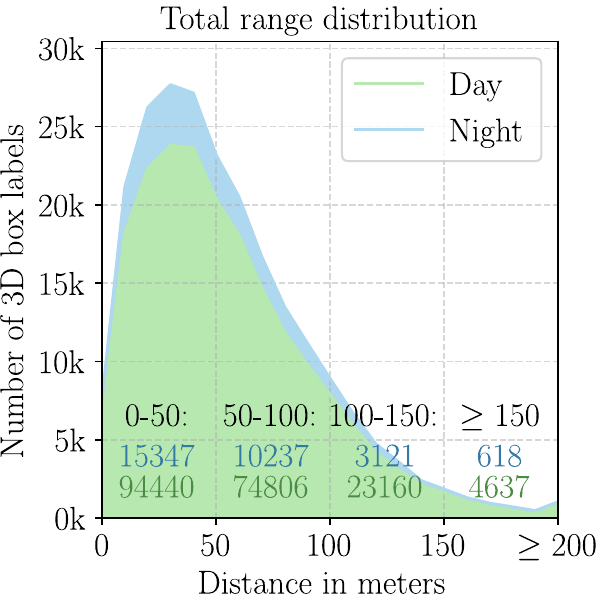}
     \end{subfigure}
     
     \caption{Range distribution per category}
     \label{fig:range_grid}
\end{figure*}

This section provides a quantitative breakdown of the WorkIng-Highway dataset across environmental conditions and data sources. As illustrated in Figure~\ref{fig:range_grid}, the majority of object labels are concentrated within the $0$--$100$\,m range. Specifically:

\noindent \textbf{Detection Range:} For the majority of classes, including Cars, Vans, and Trucks, the highest concentration of labels is located within the $0$--$100$\,m range.

\noindent \textbf{Heavy Vehicles:} Trucks and Trailers show a more distributed density beyond $100$\,m compared to smaller classes, likely a result of their larger physical profiles.

\noindent \textbf{Work Zone Assets:} Barriers show a sharp peak within the $0$--$50$\,m range, which is consistent with the placement of road-side obstacles in highway work zones.

The dataset maintains annotations across both day and night conditions, though spatial density varies:
\noindent \textbf{Diurnal Comparisons:} While the Day subset (green) provides a higher total volume, the Night subset (blue) demonstrates a comparable spatial distribution across the $0$--$200$\,m range.

\noindent \textbf{Long-Range Samples:} The dataset includes annotations beyond $150$\,m for high-visibility objects like Buses and Trucks, supporting research into long-range highway perception.

\subsubsection{Vehicle Motion Statistics}
\begin{figure*}[h]
  \centering
  \includegraphics[width=\linewidth]{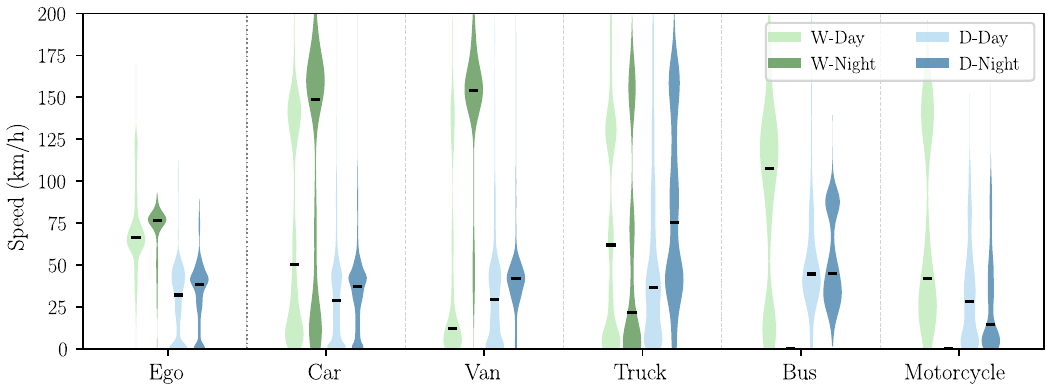}\\[3pt]
  \includegraphics[width=\linewidth]{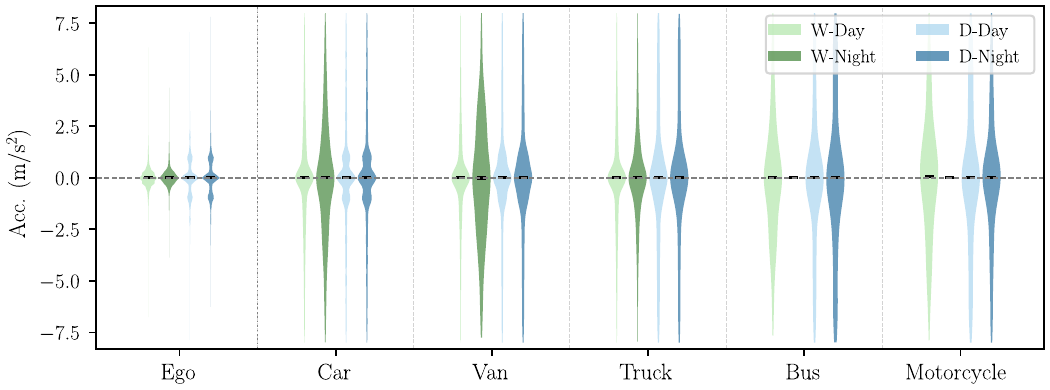}\\[3pt]
  \includegraphics[width=\linewidth]{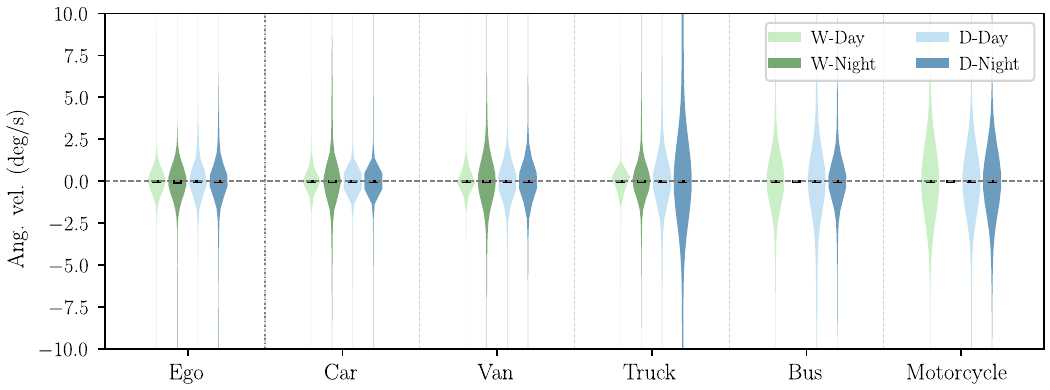}
  \caption{Vehicle motion distributions per class: speed (top),
           acceleration (middle), angular velocity (bottom).
           Light/dark green\,=\,WorkIng-Highway Day/Night;
           light/dark blue\,=\,DrivIng Day/Night.
           Black tick\,=\,median; dotted line separates ego from NPC classes.}
  \label{fig:motion_stats}
\end{figure*}

We compare vehicle motion across four recording splits: WorkIng-Highway Day/Night
and DrivIng Day/Night.
Figure~\ref{fig:motion_stats} shows the full distributions across all classes.

\noindent\textbf{Speed.}
WorkIng-Highway vehicles travel at motorway speeds: the ego drives at around
67\,km/h during the day and slightly faster at night (77\,km/h) when the road
is less busy and cruising becomes steadier.
DrivIng vehicles are noticeably slower (ego around 32--38\,km/h), typical of
mixed-road driving with more frequent speed changes.
An interesting case is the WorkIng-Highway Day Van, which only averages
12\,km/h; these are construction-support vehicles parked or moving slowly
inside the work zone.

\noindent\textbf{Acceleration.}
All vehicles show near-zero median acceleration regardless of condition,
as expected on open roads.
The key difference is how smooth the driving is: WorkIng-Highway ego
acceleration varies less (MAD\,=\,0.25\,m/s$^2$) than DrivIng
(MAD\,$\approx$\,0.55\,m/s$^2$), reflecting the steadier flow of
uninterrupted motorway driving versus the more reactive mixed-road traffic.

\noindent\textbf{Turning behaviour.}
All vehicle classes drive mostly straight: angular velocity medians are zero
across the board.
However, vehicles in WorkIng-Highway turn far less than those in DrivIng:
the 95th-percentile heading change rate is 2--4$\times$ lower for cars, vans,
and trucks (3--6\,°/s vs.\ 12--14\,°/s during the day), directly reflecting
how the narrow barrier-lined work-zone corridor limits lane changes and lateral
manoeuvres.
Motorcycles are the exception, showing the highest turning rates in both datasets
($\sim$17\,°/s at $p_{95}$), as expected given their size and manoeuvrability.

\subsection{WorkIng-CARLA}
\subsubsection{Road Work Zone Asset catalog}
A library of specialized 3D assets was integrated into the CARLA simulator. These assets are categorized into three functional groups based on their role in traffic management and environmental complexity.

Every asset is authored against the published regulation of the jurisdiction it represents. Table~\ref{tab:regional_standards} lists the sign geometry, border and background colors, channelization-device colors, and temporary lane-marking colors mandated by each of the six supported standards; these are the specifications the meshes and textures in the catalog below are built to reproduce, and the values the \texttt{region} parameter (Sec.~\ref{sec:carla}) selects between at runtime.

\begin{table}[htbp]
\centering
\caption{Regional comparison of work-zone traffic signs, barriers, and temporary road markings. In the two sign columns, the sub-labels \textit{(Shape, Border, Bg)} give each sign's shape, border color, and background (Bg) color; the last two columns give cone/barrier and temporary lane-marking colors. ``Nordic'' covers Finland, Sweden, and Iceland. In ``Europe (most)'' (Western and Southern EU/EEA countries), a \textit{yellow warning-sign background} signals a temporary hazard, whereas ``DACH'' retains the standard \textit{white warning-sign background} even in work zones.}
\label{tab:regional_standards}
\footnotesize 
\setlength{\tabcolsep}{2pt} 

\begin{tabularx}{\columnwidth}{@{} >{\raggedright\arraybackslash}p{1.2cm} >{\raggedright\arraybackslash}X >{\raggedright\arraybackslash}X >{\raggedright\arraybackslash}p{1.5cm} >{\raggedright\arraybackslash}p{1.5cm} @{}}
\toprule
\textbf{Region/ Standard}
  & \textbf{Speed Limit} \newline {\scriptsize\textit{(Shape, Border, Bg)}}
  & \textbf{Warning Sign} \newline {\scriptsize\textit{(Shape, Border, Bg)}}
  & \textbf{Cone/ Barrier} \newline {\scriptsize\textit{(Color)}}
  & \textbf{Temp. Lane} \newline {\scriptsize\textit{(Marking Color)}} \\
\midrule

\textbf{Australia}
  & Circular, Red, White \cite{austra_2018,AS1742}
  & Diamond, Black (inset), Yellow \cite{austra_2018}
  & Fluor. Orange \cite{AS1742}
  & Yellow/\allowbreak White \cite{AS1742} \\[5pt]

\textbf{China}
  & Circular, Red, White \cite{GB5768}
  & Triangle, Black, Yellow \cite{GB5768}
  & Red/\allowbreak Orange \cite{GB5768}
  & Yellow/\allowbreak Orange \cite{GB5768_4} \\[5pt]

\textbf{Nordic}
  & Circular, Red, Yellow \cite{FinnishTrafficAct_2018,SwedishRoadSign_2007,IcelandicSign_1995}
  & Triangle, Red, Yellow \cite{FinnishTrafficAct_2018,SwedishRoadSign_2007,IcelandicSign_1995}
  & Red/\allowbreak Orange \cite{TrafikverketTRVK_2023,EN1317:2010}
  & Yellow \cite{TrafikverketTRVK_2023,FinnishTrafficAct_2018,IcelandicSign_1995} \\[5pt]

\textbf{USA}
  & Rectangle (Vert.), Black/\allowbreak None, White \cite{MUTCD:2023}
  & Diamond, Black, Orange \cite{MUTCD:2023}
  & Fluor. Orange \cite{MUTCD:2023}
  & White/\allowbreak Yellow \cite{MUTCD:2023} \\[5pt]

\textbf{Europe (most)}$^{\,\ddagger}$
  & Circular, Red, White \cite{UNECE:Consolidated2006}
  & Triangle, Red, Yellow \cite{UNECE:Consolidated2006,ERF:TemporaryMarkings}
  & Red/\allowbreak Orange \cite{EN13422:2004,EN1317:2010}
  & Yellow/\allowbreak Orange \cite{ERF:TemporaryMarkings} \\[5pt]

\textbf{DACH}
  & Circular, Red, White \cite{UNECE:Consolidated2006,GermanStVO}
  & Triangle, Red, White \cite{GermanStVO}
  & Red/\allowbreak Orange \cite{EN13422:2004,EN1317:2010}
  & Yellow \cite{GermanStVO} \\

\bottomrule
\multicolumn{5}{@{} p{\columnwidth} @{}}{%
  {\scriptsize $^{\ddagger}$Covers France, Spain, Italy, Norway, Portugal, Denmark, United Kingdom, Belgium, Netherlands, Czech Republic, and other Vienna Convention
  signatories not in the DACH group.}}
\end{tabularx}
\end{table}

\begin{figure*}[htbp]
    \centering
    \includegraphics[width=\textwidth]{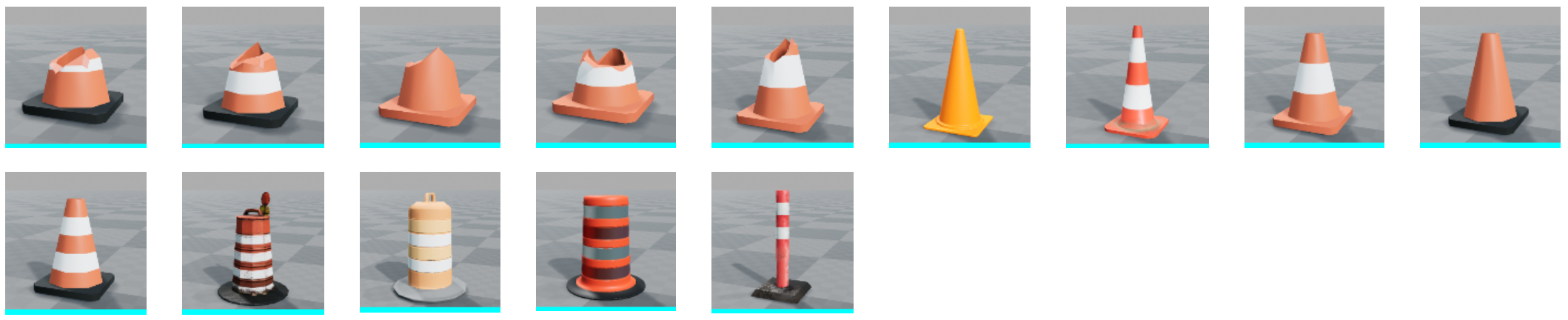}
    \caption{CARLA Asset Catalog: Traffic Cone}
    \label{fig:cone}
\end{figure*}

\begin{figure*}[htbp]
    \centering
    \includegraphics[width=\textwidth]{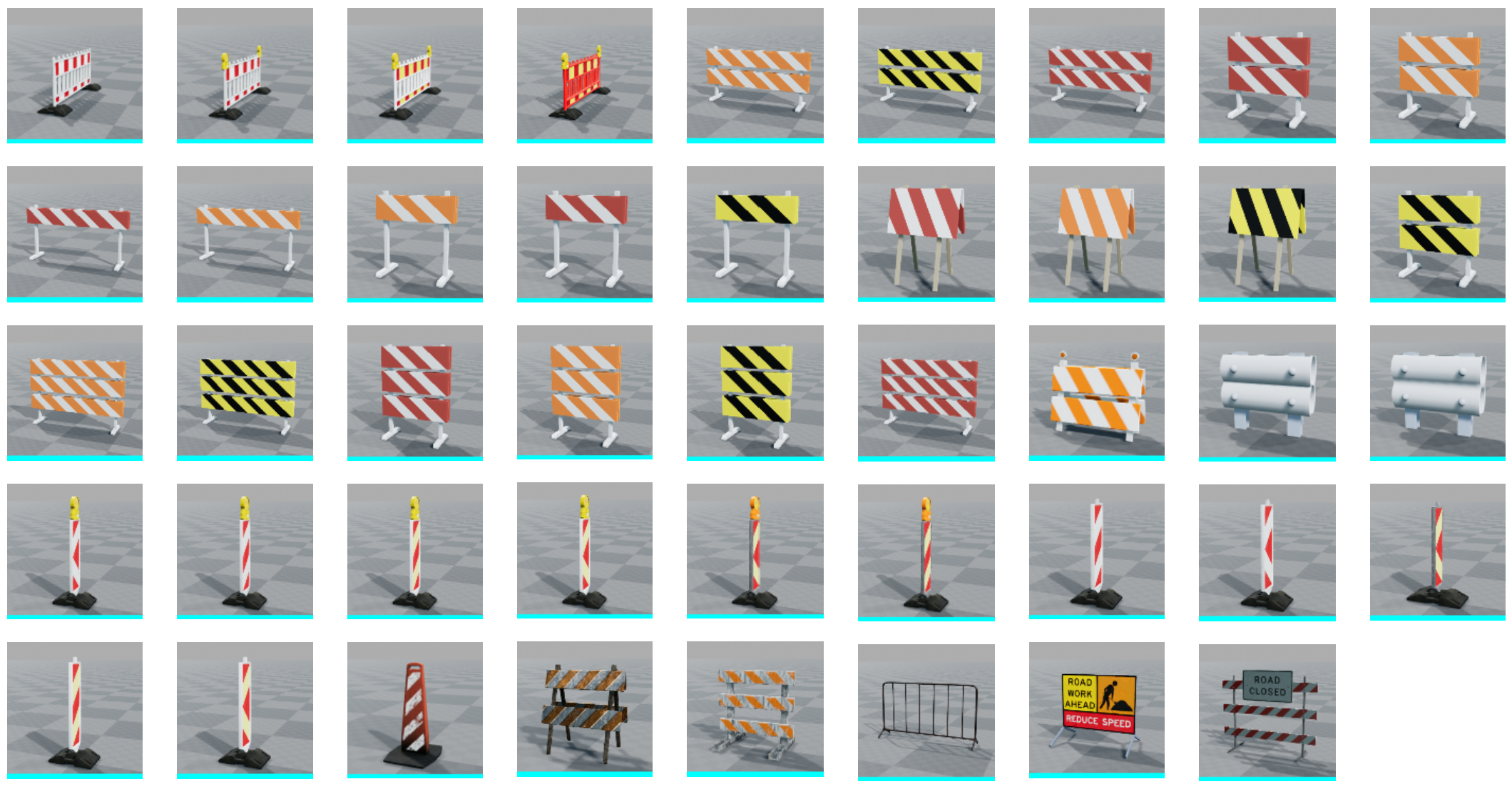}
    \caption{CARLA Asset catalog: Barrier}
    \label{fig:barrier}
\end{figure*}

\begin{figure*}[htbp]
    \centering
    \includegraphics[width=\textwidth]{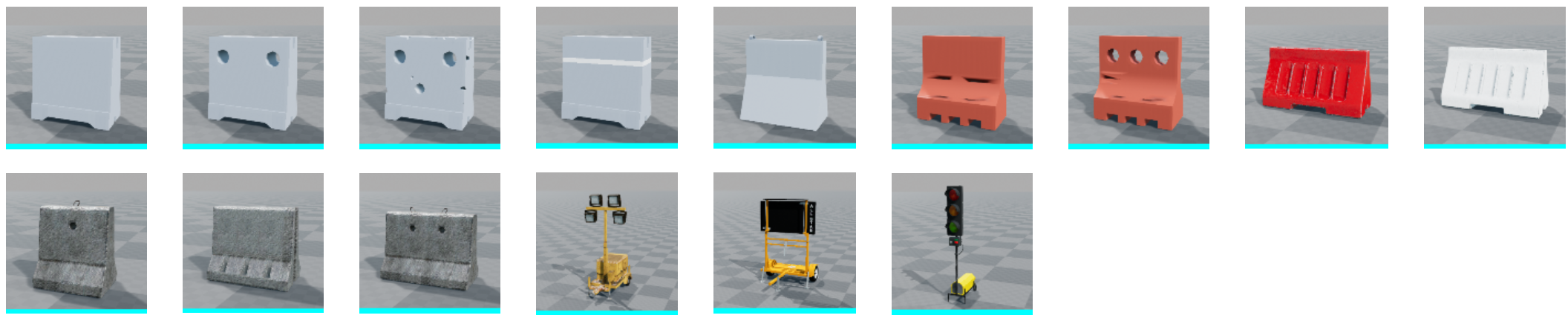}
    \caption{CARLA Asset catalog: Block}
    \label{fig:block}
\end{figure*}

\noindent \textbf{Delineation Devices and Barrier Systems} These assets define the spatial boundaries of work zones and manage longitudinal traffic flow.
\begin{itemize}
    \item \textbf{Cones:} The library contains multiple iterations of traffic cones, drums, and vertical delineators. Variations include differing heights, reflective striping patterns, and simulated physical degradation to provide geometric diversity (see Figure~\ref{fig:cone}).
    \item \textbf{Barriers \& Blocks:} These groups include modular traffic barriers, such as standard concrete "Jersey" blocks, water-fillable plastic channelizers, and modular metal fencing (see Figure~\ref{fig:barrier} and Figure~\ref{fig:block}).
\end{itemize}

\begin{figure*}[htbp]
    \centering
    \includegraphics[width=\textwidth]{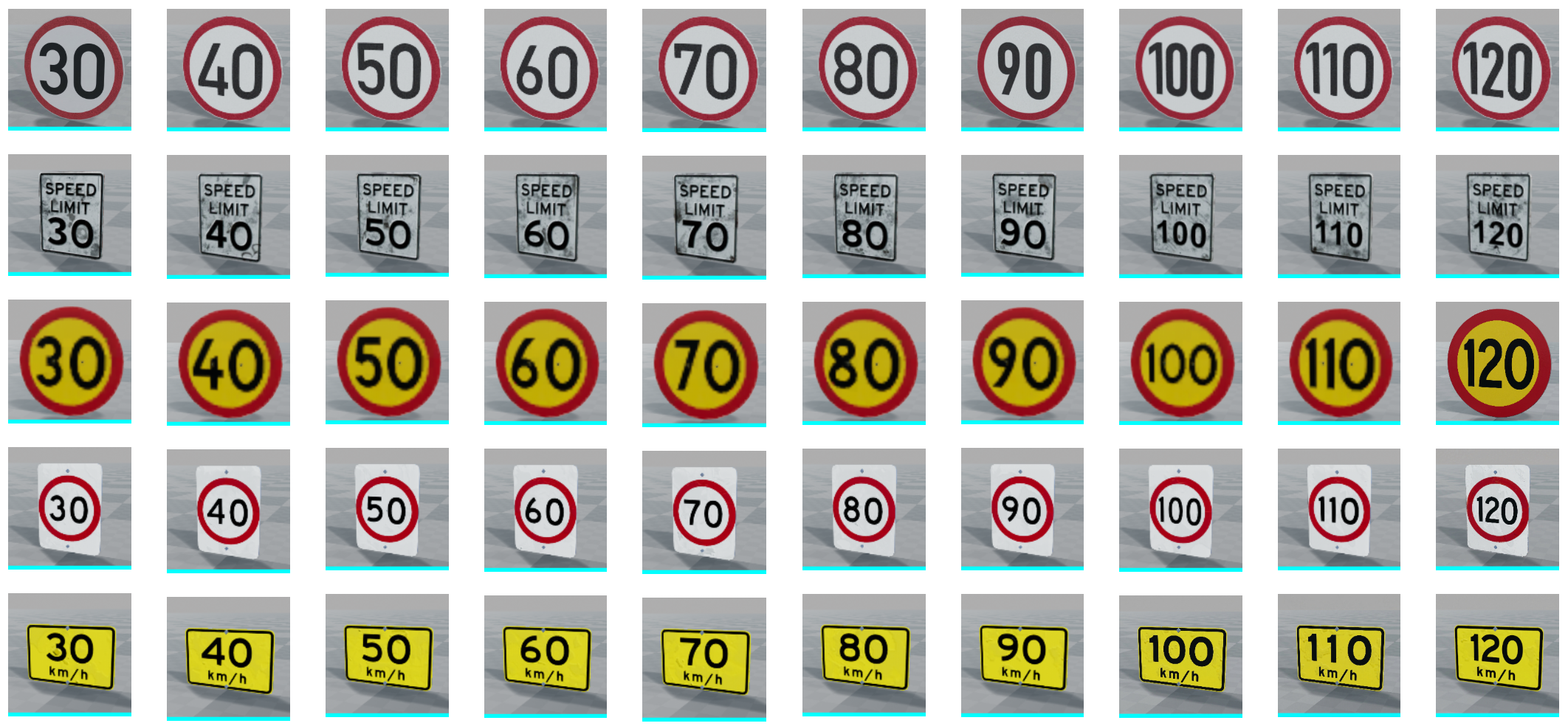}
    \caption{CARLA Asset catalog: Speed Regulatory Signage (30--120\,km/h)}
    \label{fig:speed_limit}
\end{figure*}

\begin{figure*}[htbp]
    \centering
    \includegraphics[width=\textwidth]{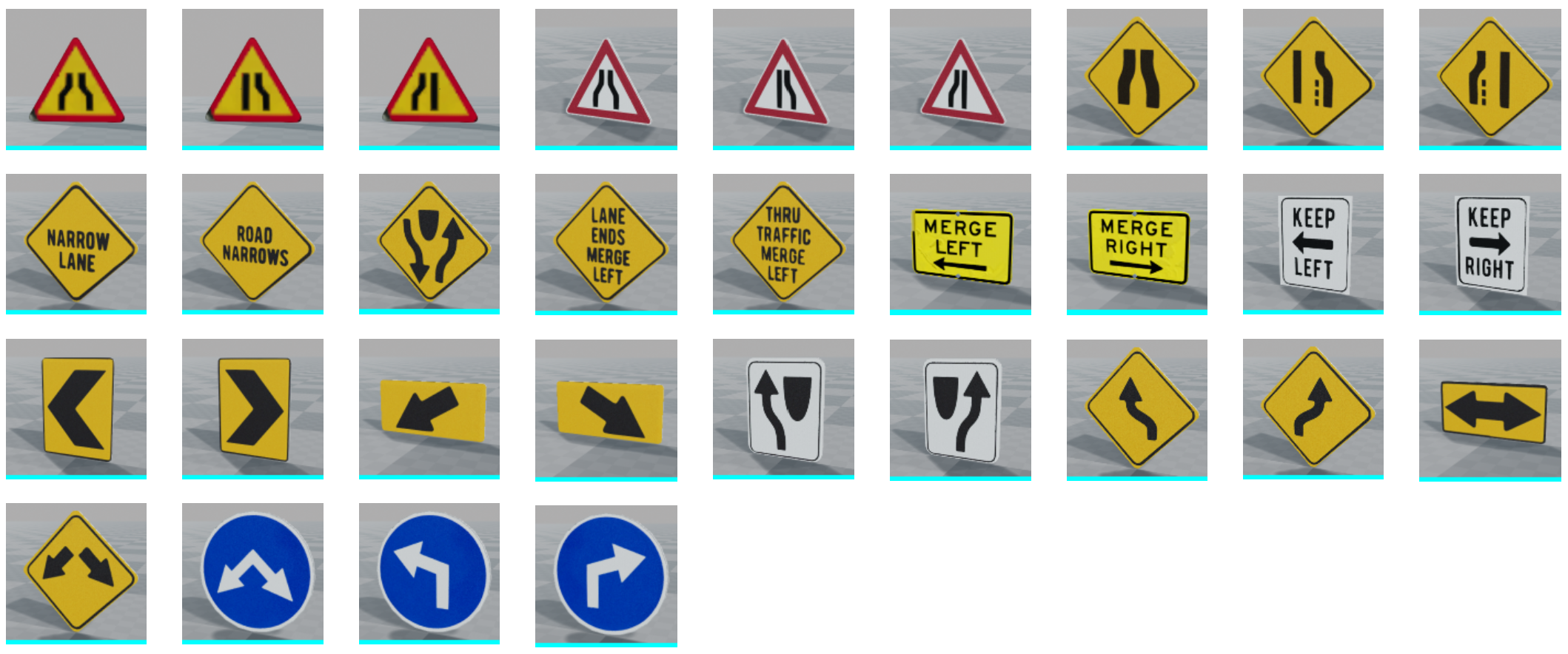}
    \caption{CARLA Asset catalog: Temporary and Fixed Warning Signage}
    \label{fig:warning_sign}
\end{figure*}

\noindent \textbf{Traffic Signage and Regulatory Devices} This category includes both permanent and temporary regulatory signage used for driver guidance through construction zones.
\begin{itemize}
    \item \textbf{Speed Limit Regulators:} Speed limit indicators cover 30--120\,km/h. The catalog includes multiple regional design variations (see Figure~\ref{fig:speed_limit}).
    \item \textbf{Warning Signs:} This subset comprises standard highway warning indicators, including "Road Narrows," "Merge Left/Right," and lane-transition symbols. These are provided in various mounting configurations, such as temporary portable stands and fixed-post installations (see Figure~\ref{fig:warning_sign}).
\end{itemize}

\begin{figure*}[htbp]
    \centering
    \includegraphics[width=\textwidth]{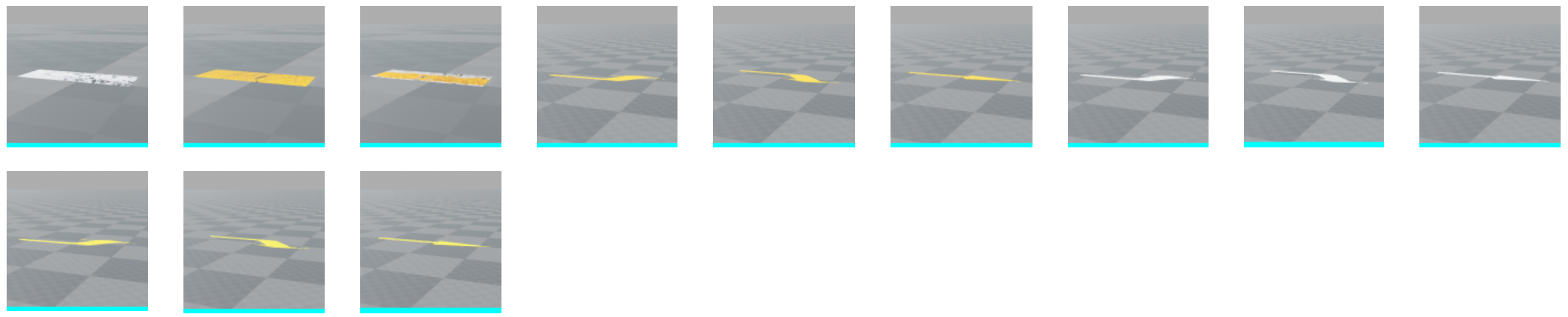}
    \caption{CARLA Asset catalog: Road Marking}
    \label{fig:marking}
\end{figure*}

\begin{figure*}[htbp]
    \centering
    \includegraphics[width=\textwidth]{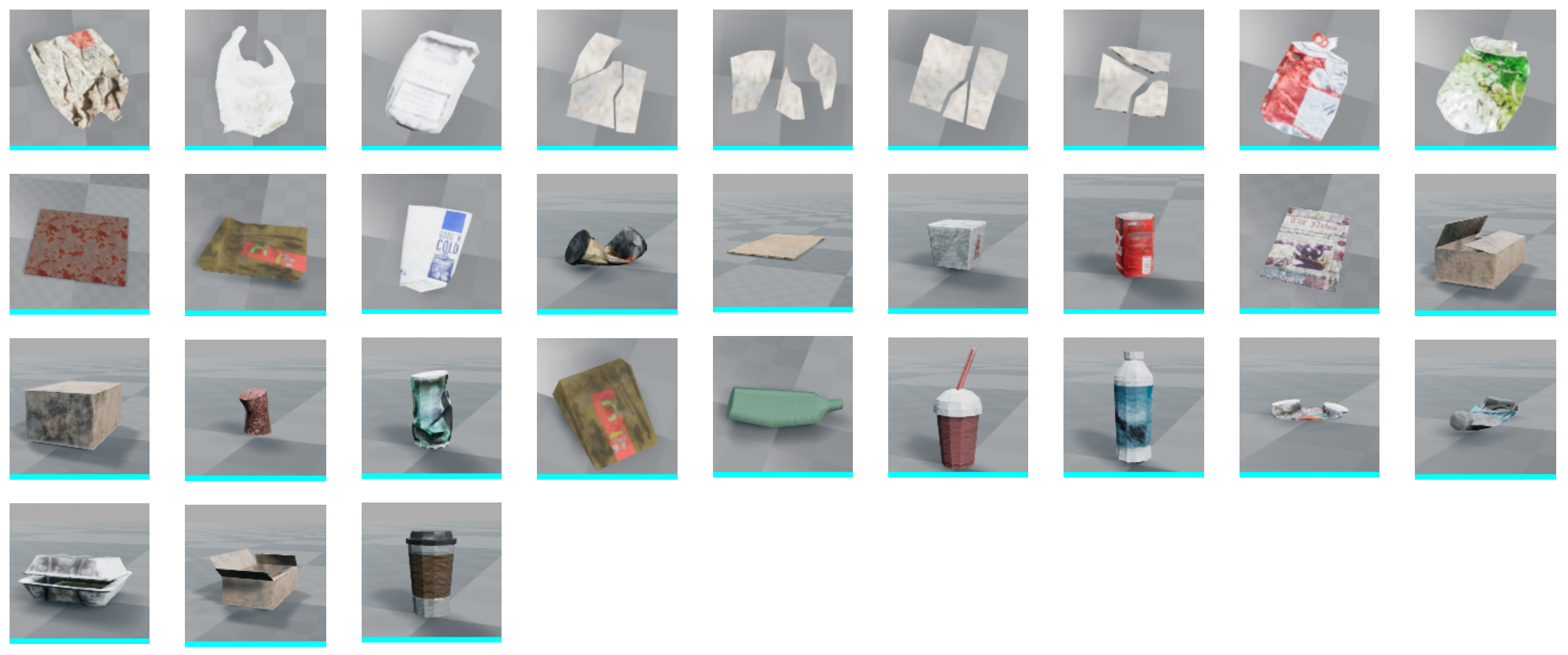}
    \caption{CARLA Asset catalog: Road Debris}
    \label{fig:debris}
\end{figure*}

\noindent \textbf{Environmental Elements and Debris} To increase the complexity of the perception environment and evaluate model robustness against non-target objects, various surface and environmental assets were developed.
\begin{itemize}
    \item \textbf{Surface Markings:} Specialized 3D meshes for horizontal lane markings were implemented, including straight segments, directional arrows, and dashed lines in white and yellow configurations (see Figure~\ref{fig:marking}).
    \item \textbf{Road Debris:} A catalog of inorganic waste materials, including discarded containers, cardboard fragments, and metallic waste (see Figure~\ref{fig:debris}).
\end{itemize}

\subsubsection{Scenario Distribution}
\begin{figure*}[htbp]
    \centering
    \includegraphics[width=\linewidth]{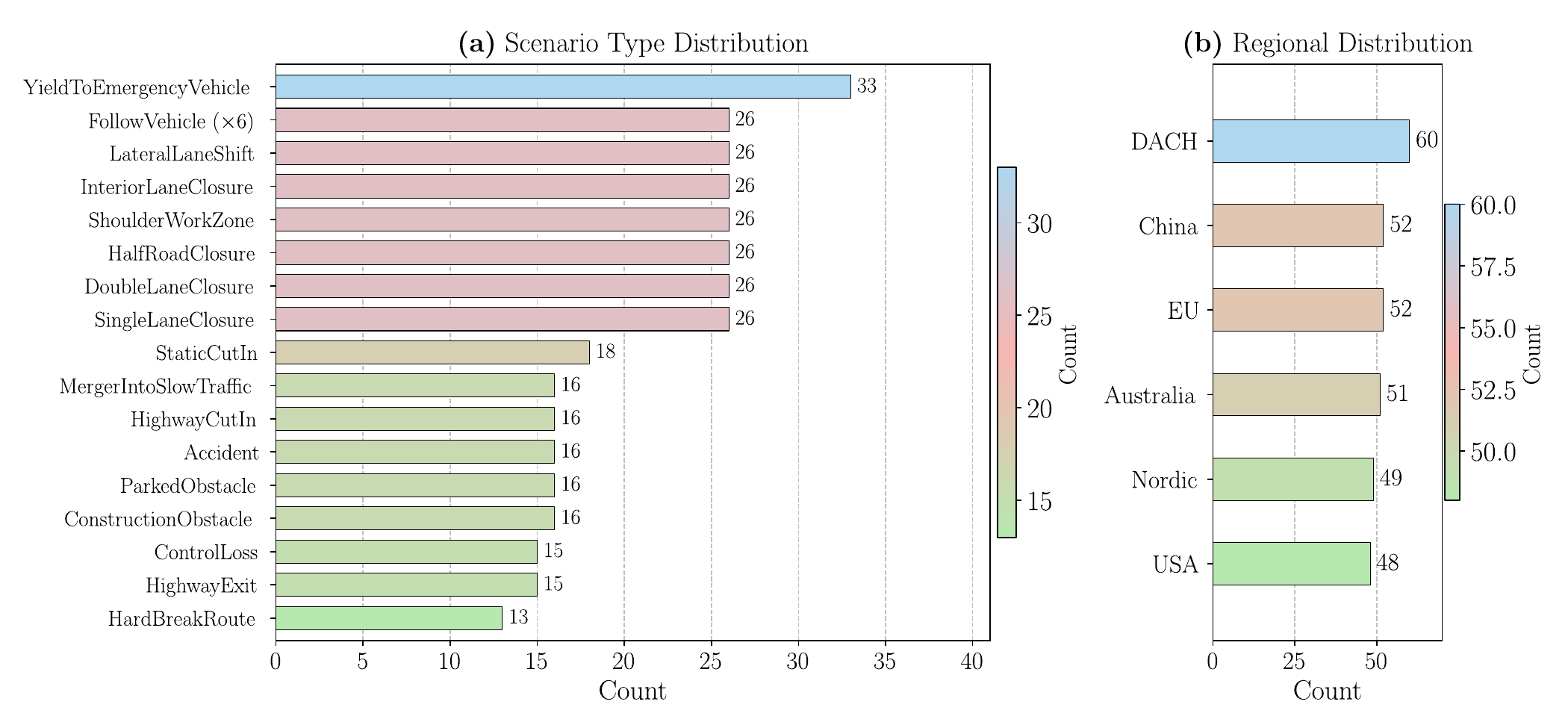}
    \caption{Quantitative characterization of WorkIng-CARLA environment. (a) Distribution of maneuver-based scenarios across the dataset. (b) Regional distribution of CARLA scenarios}
    \label{fig:scenario_statistic_1}
\end{figure*}

Figure~\ref{fig:scenario_statistic_1} characterises the 486 WorkIng-CARLA episodes: (a) their distribution over maneuver types, and (b) the regional composition of the asset library, sampled stochastically across the six supported standards. Work-zone episodes instantiate the six canonical layouts of Section~\ref{sec:carla} under both the clear and the occluded \textit{FollowVehicle} approach, in left- and right-handed mirrors. The remainder are the general highway maneuvers listed in Section~\ref{sec:carla}, included to prevent catastrophic forgetting of basic driving skills.

\subsection{Benchmark: Extended Analysis}
\label{sec:supp_detection}
This section complements the main detection benchmark (Table~\ref{tab:detection_benchmark}) with two analyses: a per-class decomposition of the aggregate mAP reported in the main paper, and a qualitative study of representative cases analysis for each modality.

\subsubsection{Per-Class mAP Breakdown}
Table~\ref{tab:supp_detection_breakdown} decomposes the aggregate mAP from the main benchmark into per-category Average Precision for all three baselines under Day, Night, and CARLA conditions. The breakdown substantiates the modality trends discussed in Section~\ref{sec:result_detection}: the camera-only baseline (PETR) degrades most severely on small and appearance-dependent classes, most notably Pedestrian ($0.12$ Day) and Bus ($0.18$ Day), where monocular depth estimation is least reliable, whereas the LiDAR-based detectors (CenterPoint, BEVFusion) remain comparatively robust on large, rigid actors and barriers. The Night columns also make the evaluation-scope effect explicit: with Bus, Motorcycle, and Pedestrian excluded due to insufficient annotation density, the aggregate concentrates on larger, more reliably detected classes, which mathematically inflates the Night totals relative to Day.

\begin{table}[t]
\centering
\caption{\textbf{Class-specific 3D object detection performance.} Detailed mean Average Precision (mAP) breakdown across all evaluated object categories under Day, Night, and CARLA simulation conditions. Entries marked with “-” denote classes excluded from evaluation in that scenario, either absent from the split or below the annotation-density threshold (Section~\ref{sec:benchmarking}). Bold values denote the best performance in each scenario.}
\label{tab:supp_detection_breakdown}
\scriptsize
\setlength{\tabcolsep}{2pt} 
\begin{tabularx}{\columnwidth}{@{} l >{\raggedright\arraybackslash}p{1.7cm} *{8}{>{\centering\arraybackslash}X} @{}}
\toprule
\textbf{Scene} & \textbf{Model} & \textbf{Car} & \textbf{Tru.} & \textbf{Bus} & \textbf{Trai.} & \textbf{Bar.} & \textbf{Mot.} & \textbf{Ped.} & \textbf{Total} \\ \midrule

\multirow{3}{*}{Day}
  & CenterPoint \cite{yin2021center}   & \textbf{0.95} & \textbf{0.89} & \textbf{0.37} & \textbf{0.87} & \textbf{0.86} & 0.68          & \textbf{0.71} & \textbf{0.76}          \\
  & PETR \cite{liu2022petr}          & 0.60          & 0.47          & 0.18          & 0.39          & 0.39          & 0.45          & 0.12          & 0.37          \\
  & BEVFusion \cite{liu2022bevfusion} & \textbf{0.95} & 0.85          & \textbf{0.37} & 0.83          & 0.80          & \textbf{0.80} & 0.64          & 0.75 \\ \midrule

\multirow{3}{*}{Night}
  & CenterPoint \cite{yin2021center}   & 0.88          & \textbf{0.87} & -             & \textbf{0.44} & \textbf{0.92} & -             & -             & \textbf{0.78} \\
  & PETR \cite{liu2022petr}          & 0.43          & 0.57          & -             & 0.01          & 0.30          & -             & -             & 0.33          \\
  & BEVFusion \cite{liu2022bevfusion} & \textbf{0.93} & 0.85          & -             & 0.42          & 0.90          & -             & -             & \textbf{0.78} \\ \midrule

\multirow{3}{*}{CARLA}
  & CenterPoint \cite{yin2021center}   & \textbf{0.99} & \textbf{0.98} & \textbf{0.99} & -             & \textbf{0.88} & \textbf{0.98} & \textbf{0.98} & \textbf{0.97} \\
  & PETR \cite{liu2022petr}          & 0.64          & 0.52          & 0.53          & -             & 0.65          & 0.55          & 0.49          & 0.56          \\
  & BEVFusion \cite{liu2022bevfusion} & 0.97          & 0.95          & 0.94          & -             & 0.82          & 0.96          & 0.92          & 0.93          \\ \bottomrule
\end{tabularx}
\end{table}

\subsubsection{Qualitative Results and Failure Case Analysis}
\textbf{Failure Analysis Methodology} To systematically characterise model errors beyond aggregate mAP,
we adopt a TIDE-style~\cite{bolya2020tide} error decomposition
adapted to 3D detection with nuScenes centre-distance matching.
Each prediction above the score threshold is assigned exactly one
error type: \textbf{TP} (matched, correct class, $d{<}2$\,m),
\textbf{Cls} (matched but wrong class), \textbf{Loc} (unmatched
but a same-class GT exists within 4\,m), or \textbf{Bkg}
(spurious false positive).  Unmatched ground-truth instances are
labelled \textbf{Miss}.  Performance is further stratified by
distance (0--20\,m, 20--40\,m, 40--60\,m) and point density
(sparse $\leq$5\,pts, medium, dense) to isolate range and
occlusion effects.

\textbf{Cross-model comparison.}
For each evaluation token we compute per-model F1 and report the
\emph{F1-spread} (max\,$-$\,min across models).  Tokens are
categorised as \emph{model-diverge} (spread${>}0.4$, one model
succeeds where another fails), \emph{all-good} (all
F1${>}0.7$), or \emph{all-bad} (all F1${<}0.3$).  This surface
reveals whether failure is model-agnostic (data difficulty) or
architecture-specific (a tractable engineering gap).

We refer the reviewer to the supplemental material (external PDF) for per-model visual breakdowns and a curated gallery of representative success and failure cases across all three architectures.

\subsection{2D Work-Zone Asset Detection: Method and Extended Results}
\label{supp:2d_asset_detection}

This section provides the full method and extended results for the complementary 2D asset-recognition ablation summarized in Sec.~\ref{sec:ablation_2d}.

\subsubsection{Motivation}
Beyond 3D perception, the visual diversity of work-zone signage and channelization furniture is itself a distinct challenge: unlike general-purpose object classes, work-zone assets are small, visually similar within class (e.g., orange cones vs.\ orange barrels), frequently partially occluded by other furniture or vehicles, and vary systematically in shape, color, and text across jurisdictions (Table~\ref{tab:regional_standards}). A natural question raised by this diversity is whether WorkIng-CARLA's procedurally-generated assets are visually and geometrically realistic enough to serve as useful training signal for 2D recognition models. We therefore export WorkIng-CARLA's 3D work-zone annotations to 2D bounding boxes and instance segmentation masks under ROADWork's~\cite{ghosh2025roadwork} 16-category taxonomy expect \textit{Work Vehicle}, and evaluate whether training on this synthetic data, alone and combined with real ROADWork imagery, improves recognition of real-world work-zone assets.

\subsubsection{Method: 2D Annotation Pipeline}
We export axis-aligned 2D boxes and instance masks for every annotated 3D object in WorkIng-CARLA, reusing the 3D ground truth of the detection benchmark (Sec.~\ref{sec:carla}) and projecting onto the front-left and front-right corner cameras (Figure~\ref{fig:sensor_placement}), the closest match to ROADWork's front-facing monocular setup. Masks are taken from CARLA's instance-segmentation camera, matched to each projected 3D box and filtered by an occlusion check; the full procedure is in the released code. Blueprint identifiers are mapped onto ROADWork's 16-category taxonomy, for which all but \textit{Work Vehicle} have a corresponding asset, from our custom library (Table~\ref{tab:asset_catalog}) and official CARLA blueprints for the worker and police classes~\cite{carla_vehicles}.

\subsubsection{Extended Discussion}
Table~\ref{tab:ablation_2d} (Section~\ref{sec:ablation_2d}) reports the full metric suite, including the COCO small/medium/large breakdown and the three zero-shot open-vocabulary baselines (OpenSeeD~\cite{zhang2023openseed}, Detic~\cite{zhou2022detic}, OMG-LLaVA~\cite{zhang2024omgllava}). This section covers what the main text does not: why the zero-shot baselines fail, what limits the synthetic-only configuration, and where the \textbf{C+R} gain is concentrated.

\textbf{Zero-shot foundation models fail on work-zone asset recognition.} All three score below $4.2$ AP, and OMG-LLaVA fails entirely ($0.2$ AP). This matches ROADWork's own finding that open-vocabulary methods, despite web-scale pretraining, do not transfer to the specific, regulation-defined appearance of work-zone furniture and signage without domain-specific fine-tuning. The comparison against \textbf{C}-only training in Sec.~\ref{sec:ablation_2d} should be read with one caveat: the zero-shot models receive no fine-tuning on this taxonomy, so the margin measures the value of task-specific supervision and is not a like-for-like architecture comparison.

\textbf{The residual sim-to-real gap for 2D recognition is a catalog limitation, not a rendering-fidelity one.} Unlike the 3D detection ablation, where the sim-only gap is commonly attributed to photometric domain shift, the 2D sim-only gap here has two identifiable structural causes, at different levels of the taxonomy. At the category level, the catalog covers every ROADWork class except \textit{Work Vehicle}, so \textbf{C}-only models cannot predict that category under any circumstance. Within the categories it does cover, the traffic-sign templates reproduce the regulation-defined designs of each jurisdiction (Table~\ref{tab:regional_standards}) but not the long tail of non-standardized, site-specific signage that ROADWork's own annotations document extensively. Both are catalog-completeness limitations, not rendering-fidelity ones, and both are addressable by extending the asset library.

\textbf{The \textbf{C+R} gain is concentrated in small and medium objects.} The relative AP improvement reported in Sec.~\ref{sec:ablation_2d} exceeds the relative NDS gains for 3D detection (Table~\ref{tab:ablation_training}). It concentrates in AP\textsubscript{S} ($+14.6\%$ relative for Mask R-CNN) and AP\textsubscript{M}, where WorkIng-CARLA's dense per-scenario asset placement supplies exactly the small, repetitive channelization furniture (cones, barrels, delineators) that is comparatively rare in ROADWork's real imagery. AP\textsubscript{L} is the one exception, decreasing slightly for both architectures ($-7.5\%$ and $-7.3\%$ relative): large ROADWork instances are disproportionately Work Vehicle, the one class the catalog does not contain and therefore the only category the synthetic partition cannot reinforce. This confirms the catalog-completeness explanation: the synthetic partition helps where its asset coverage is strong and is neutral-to-negative where it is absent.

\end{document}